\documentclass[10pt,twocolumn,letterpaper]{article}

\usepackage{cvpr}
\usepackage{times}
\usepackage{epsfig}
\usepackage{graphicx}
\usepackage{amsmath}
\usepackage{amssymb}
\usepackage{algorithm, algpseudocode}
\usepackage[draft]{minted}
\usepackage{multirow}
\usepackage{booktabs}
\usepackage{subcaption}
\usepackage{float}
\usepackage[dvipsnames]{xcolor}
\usepackage[accsupp]{axessibility} 

\definecolor{identifiercolor}{rgb}{.4,.6,.56}
\definecolor{stringcolor}{gray}{0.5}
\definecolor{inactivecolor}{rgb}{0.15,0.15,0.5}
\usepackage{listings}
\usepackage{mathtools}

\DeclarePairedDelimiter\floor{\lfloor}{\rfloor}

\algnewcommand{\algorithmicforeach}{\textbf{for each}}
\algdef{SE}[FOR]{ForEach}{EndForEach}[1]
  {\algorithmicforeach\ #1\ \algorithmicdo}
  {\algorithmicend\ \algorithmicforeach}

\newcommand{\ddim}{\mbox{dim}}
\newcommand{\conv}{\mbox{Conv}}
\newcommand{\convT}{\mbox{ConvT}}
\newcommand{\lin}{\mbox{Linear}}
\newcommand{\resh}{\mbox{Reshape}}

\usepackage[pagebackref=true,breaklinks=true,letterpaper=true,colorlinks,bookmarks=false]{hyperref}

\begin{document}

\title{Sampling From Autoencoders’ Latent Space via Quantization
And Probability Mass Function Concepts}

\author{Aymene Mohammed Bouayed $^{\dagger \, \ddagger}$, Adrian Iaccovelli $^{\ddagger}$, David Naccache $^{\dagger}$\\
$^{\dagger}$DIÉNS, ÉNS, CNRS, PSL University, Paris, France\\
45 rue d’Ulm, 75230, Paris cedex 05, France \\ 
{\tt\small firstname.lastname@ens.fr}\\
$^{\ddagger}$Be-Ys Reseach, France\\
46 rue du ressort, Clermont-Ferrand, France\\
{\tt\small firstname.lastname@be-ys-research.com}
}


\maketitle
\thispagestyle{empty}

\begin{abstract}
\vspace{-.3cm}In this study, we focus on sampling from the latent space of generative models built upon autoencoders so as the reconstructed samples are lifelike images. To do to, we introduce a novel post-training sampling algorithm rooted in the concept of probability mass functions, coupled with a quantization process.
Our proposed algorithm establishes a vicinity around each latent vector from the input data and then proceeds to draw samples from these defined neighborhoods. This strategic approach ensures that the sampled latent vectors predominantly inhabit high-probability regions, which, in turn, can be effectively transformed into authentic real-world images. A noteworthy point of comparison for our sampling algorithm is the sampling technique based on Gaussian mixture models (GMM), owing to its inherent capability to represent clusters. Remarkably, we manage to improve the time complexity from the previous $\mathcal{O}(n\times d \times k \times i)$ associated with GMM sampling to a much more streamlined $\mathcal{O}(n\times d)$, thereby resulting in substantial speedup during runtime. Moreover, our experimental results, gauged through the Fréchet inception distance (FID) for image generation, underscore the superior performance of our sampling algorithm across a diverse range of models and datasets.
On the MNIST benchmark dataset, our approach outperforms GMM sampling by yielding a noteworthy improvement of up to $0.89$ in FID value. Furthermore, when it comes to generating images of faces and ocular images, our approach showcases substantial enhancements with FID improvements of $1.69$ and $0.87$ respectively, as compared to GMM sampling, as evidenced on the CelebA and MOBIUS datasets.
Lastly, we substantiate our methodology's efficacy in estimating latent space distributions in contrast to GMM sampling, particularly through the lens of the Wasserstein distance. 
  
\end{abstract}

\section{Introduction}
\vspace{-.1cm}The realm of image generation in biometric imaging has undergone a notable surge in recent times, with a variety of generative models rooted in Generative Adversarial Networks (GANs) \cite{GANs}, as in \cite{iris_gan,identity_leakage_GAN, 4D_cycle_GAN}. In contrast, our approach charts a distinct path by employing autoencoder-based generative models for image synthesis. This decision is underpinned by the intricate and potentially unstable learning phase inherent to GANs, where improper tuning can lead to complications such as mode collapse \cite{Collapse_mode_GAN}. As a result, our focus in this study centers on the nuanced task of sampling within the latent space of autoencoders.

Existing literature presents three primary strategies for latent space sampling within autoencoders. The first approach involves the incorporation of a prior distribution onto the latent space through a regularization term. To acquire a vector sample, we draw from this prior distribution. Variational Autoencoders (VAEs) \cite{VAE} are a representative model category following this strategy. However, a limitation of these methods emerges after training, where the latent space might not precisely adhere to the predefined prior distribution, leading to the emergence of clusters within the latent space \cite{gmm-10}. Consequently, samples drawn from the prior distribution might deviate from the actual ground truth distribution. Consequently, the decoder model struggles to faithfully reconstruct these samples into tangible real-world data points \cite{gmm-10}.

The second avenue revolves around modeling clusters within the latent space through a Gaussian mixture model (GMM)-based sampling approach. This method entails fitting a GMM to the latent vectors from the training data, then generating new samples from this learned mixture model~\cite{gmm-10}. Generally, this technique yields superior samples compared to sampling from a prior. However, due to the expansive nature of the Gaussian distribution across the entire Euclidean space, there are instances where generated samples fall outside the decoder network's capacity to transform them into authentic real-world samples~\cite{chadebec_data_2021}.

Lastly, an alternative strategy involves the random sampling of latent vectors, followed by their manipulation through either discrete or continuous normalizing \mbox{flows~\cite{HVAE, flow_model_2}}, aiming to guide them towards high-density regions. Various methodologies to identify such high-density regions have been formulated in the literature \cite{HVAE, chadebec_data_2021, flow_model_2,vae_discrete_flows}. Nonetheless, these techniques entail significant computational overhead and are applicable only to specific types of autoencoder-based models necessitating tailored model architectures \cite{NF}. For instance, the approach presented in \cite{chadebec_data_2021} introduces supplementary learnable layers to estimate the metric of the latent space, which assumes the form of a Riemannian manifold. 

Within the scope of this study, we craft a novel technique for sampling from latent spaces, a methodology that can be seamlessly integrated with any autoencoder model. Our approach effectively sidesteps the challenge of drawing samples from regions that cannot be faithfully reconstructed into authentic real-world images, a shortcoming often encountered in GMM sampling. This bears significance in biometrics, where superior latent samples can be adeptly transformed into highly realistic synthetic images.
As a solution, we present a post-training density estimation algorithm designed to operate on the latent space of \textsl{any} autoencoder-based generative model. This algorithm employs \textsl{both a quantization step and the concept of probability mass function}. The salient attribute of our algorithm lies in its impressive time complexity of $\mathcal{O}(n\times d)$, where $n$ denotes the dataset size and $d$ symbolizes the latent space dimension. This translates into a significant acceleration in runtime compared to GMM sampling. Moreover, we validate the quality of the generated images through our methodology both qualitatively and quantitatively. This validation is performed on three distinct image generation benchmark datasets (including two in the biometric domain) and five distinct autoencoder-based models.
In terms of image quality gauged by the Fréchet Inception Distance (FID) metric, our approach demonstrates a considerable enhancement compared to GMM sampling. Specifically, on the MNIST \cite{mnist}, CelebA \cite{celeba}, and \mbox{MOBIUS \cite{rot2018deep,rot2020deep,vitek2023exploring, ssbc2020, vitek2020comprehensive}} datasets, we achieve improvements of up to $0.89$, $1.69$, and $0.87$ respectively when compared to GMM sampling.

To delve into the specifics, Section \ref{related works} presents related works addressing the challenge of latent space sampling within autoencoders. In Section \ref{background}, we introduce essential background definitions and notations. The mechanics of our sampling methodology, anchored in quantization and probability mass function concepts, are presented in Section \ref{contribution}. The efficacy of our method in image generation is rigorously tested and verified across diverse datasets in Section \ref{exp}. Lastly, we conclude and outline avenues for future exploration in Section \ref{conclusion}.

\section{Related works}
\label{related works}
\subsection{Sampling from a prior}
Numerous studies have introduced a regularization term to encourage the distribution of latent vectors to approximate a predefined prior distribution, often taking the form of a normal distribution. Within this category of methods, notables include variational autoencoders (VAEs) \cite{VAE}, and \mbox{$\beta$-VAE~\cite{beta-vae}}. These approaches model the latent space as a distribution and employ the Kullback-Leibler divergence~\cite{kl_divergence} to steer each distribution towards a normal distribution. Furthermore, the landscape includes Wasserstein autoencoders (WAE) \cite{wae}, which retain the deterministic essence of autoencoders while harnessing the Wasserstein distance \cite{OP-book-Gabriel} to drive the overall distribution of latent vectors towards a Gaussian distribution.

Following the minimization of these regularization terms, the process of latent vector sampling often involves drawing from the prior distribution and subsequently reconstructing the sampled vector into a data point. Consequently, due to the inherent non-zero nature of the resulting loss after enforcing such constraints, clusters may materialize within the latent space. Additionally, the latent vectors might not precisely adhere to the predefined prior distribution \cite{gmm-10}. To counteract this challenge, $\beta$-VAEs adopt an elevated weighting for the regularization term, striving to align the distribution of latent vectors with the prior distribution. However, this intensified weighting negatively impacts the autoencoder model's reconstruction prowess and significantly hampers its generative potential \cite{beta-vae}.

\subsection{Sampling from a Gaussian mixture model}
Gaussian Mixture Models (GMMs) assume the presence of subpopulations in the data and have demonstrated high performance in multiple works \cite{pythae, gmm-10}. Therefore, GMMs model each subpopulation \(i\) via a Gaussian distribution \(\mathcal{N}_i(\cdot\vert\mu_i, \Sigma_i)\) with a mean parameter \(\mu_i\) and covariance parameter \(\Sigma_i\). To express the probability distribution of the entire data set, a weighted sum of the probability distributions \(\mathcal{N}_i(\cdot\vert\mu_i, \Sigma_i)\) by weights \(\alpha_i\) is taken, where \(\sum_i \alpha_i =1\). This way, the probability of any data point in the dataset can be written as:
\[
p(x) = \sum_i \alpha_i \mathcal{N}_i(x\vert\mu_i, \Sigma_i).
\]
The parameters of the GMM can be estimated by using the Expectation-Maximization (EM) algorithm \cite{EM}. Nonetheless, this sampling method presents two drawbacks. 1) The large value of \(\mathcal{O}(n\times d \times k \times i)\) computational complexity of the EM algorithm, where \(n, d, k\) and \(i\) represent respectively the size of the data set, the dimension of the latent space, the number of distributions in the GMM model and the number of iterations. In fact, we could have an \textsl{infinite} time complexity for this algorithm, if not for the number of iterations. 2) With low probability, outliers can be sampled, since the model is based on distributions defined on the whole latent space.

\subsection{Sampling using normalizing flows}
Recent works like in \cite{HVAE,chadebec2022AGP,chadebec_data_2021,ae_flow,vae_discrete_flows} propose to use normalizing flows \cite{NF_intro,NF} to sample latent vectors from the latent space of autoencoders. These methods select vectors randomly as sample vectors, then move them to high probability regions. Normalizing flow methods can be divided into two families, notably discrete and continuous normalizing flows \cite{NF}. The difference comes from the definition of the flow process \(T(z)\) as discrete steps via the composition of functions \(\{T_i\}^k_{i=1}\) as in \cite{ae_flow,vae_discrete_flows}:
\[
T(z) = T_k \circ \dots \circ T_1(z)
\]
or continuous via an integral over an interval as in \cite{chadebec2022AGP}:
\[
T(z)=z+\int_{t=t_0}^{t_1} g_\phi(t,z_t)dt.
\]
Recent works such as \cite{HVAE,chadebec2022AGP,chadebec_data_2021} focus mostly on continuous normalizing flows due to their better modeling of the flow dynamics. These works propose to see the Euclidean latent space of VAEs as a Riemannian manifold \cite{pennec} with the sum of all covariance matrices weighted by the distance to the mean as the manifold's metrics \cite{chadebec2022AGP}. This metric was further improved in \cite{chadebec_data_2021} by learning a covariance matrix from the data. Then, using the determinant of this metric, the probability density at any point can be estimated. Finally, via a Hamiltonian Monte Carlo process, the drawn samples are moved to regions of high probability. These regions can be reconstructed into realistic data points \cite{chadebec_data_2021}. Yet, these methods do not represent a general off-the-shelf method to sample from the latent space because they require a dedicated model architecture. Also, as noted in \cite{NF}, using Riemannian metrics is only applicable to topologies homeomorphic to Euclidean space \(\mathbb{R}^d\) which is not applicable, for example, to spherical latent spaces such as in Hyperspherical Variational Auto-Encoders \cite{SVAE}.

In this work, our proposed sampling method performs density estimation over the latent space via probability mass function and quantization processes. Consequently, it does not require a special regularization term (as in sampling from a prior distribution) nor a special autoencoder architecture (as in sampling using normalizing flows). Furthermore, in addition to not sampling outliers, training our sampling method improves the time complexity of Gaussian mixture model sampling.

\section{Background}
\label{background}
In this section, we introduce necessary background definitions and notations.

\subsection{Autoencoders}

A \textsl{Vanilla Autoencoder} \cite{Goodfellow-et-al-2016} is a neural network architecture comprised of two main components: an \textsl{encoder}~$E_\phi$ and a \textsl{decoder} $D_\theta$. The role of the encoder is to map input data $x$ from the space $\mathcal{X}$ to a latent representation $z$ within a lower-dimensional space~$\mathcal{Z}$, where $\ddim(\mathcal{Z}) \ll \ddim(\mathcal{X})$. Conversely, the decoder's function is to reconstruct the input data $\hat{x}$ using the latent representation~$z$. Both the encoder and decoder are equipped with parameters $\phi$ and $\theta$ correspondingly. The optimization of the weights $\phi$ and $\theta$ is achieved through gradient descent, aimed at minimizing the reconstruction loss. This loss is generally defined as the squared $\ell_2$ norm of the difference between the input data $x$ and its corresponding reconstruction $\hat{x}$, and can be mathematically articulated as $\mathcal{L}_{AE}(x, \hat{x}) = \vert\vert x - \hat{x}\vert\vert_2^{2}$.

\subsection{Quantization}
Quantization involves the process of discretizing an input, often derived from a continuous or extensive set of values, into a discrete integer set. In the case of a vector, each dimension is commonly treated independently.
Machine learning and autoencoder studies on quantization can be divided into two main categories: quantization-aware training and post-training quantization. In the quantization-aware training approach, quantization is applied during the learning phase. This approach yields a model that has learned to handle quantized vectors, making the resulting sampling techniques model-specific. Notable among such methods is VQ-VAE \cite{VQVAE}, which develops a dictionary of quantized vectors and maps each latent vector of input data to the nearest quantized vector. On the other hand, the second approach of post-training quantization applies quantization subsequent to the model's training. This method's advantage lies in its applicability to various models without altering the learning phase nor necessitating extra layers for fine-tuning. Within the realm of post-training quantization methods, examples include lattice quantization \cite{best_lattice_quantization, lattice_quantization} and clustering-based non-uniform quantization \cite{han2016deep, Stock2020And}. While these methods offer superior quantization quality, they tend to be computationally intensive.
In this study, we adopt a post-training quantization strategy to leverage its broad applicability. However, given our focus on simplicity and reduced computational complexity, we introduce a quantization method in Section \ref{contribution} with a time complexity of just $\mathcal{O}(d)$, where $d$ represents the length of the vector.

\subsection{Probability density function}
Given a continuous random variable $x$ taking on an innumerable infinite number of possible values with support in $\mathcal{C}$. A \textsl{probability density function} $PDF(\cdot)$ estimates the probability of $a<x<b$, i.e. $P(a<x<b)$ and satisfies the following conditions :
\begin{itemize}
    \item $PDF(x)\geq 0 \quad \forall x \in \mathcal{C}$.
    \item $\int_\mathcal{C} PDF(x) dx = 1$.
    \item $P(a<x<b) = \int_a^b PDF(x) dx$.
    \item $P(x=a) = 0 \quad \forall a \in \mathcal{C}$.
\end{itemize}

\subsection{Probability mass function}
A \emph{Probability mass function} (PMF) is an adaptation of the probability density function to the case of discrete random variables. Given a discrete random variable $x$ taking a finite or countably infinite number of possible values with support in $\mathcal{D}$, a PMF estimates the probability $P(x = a) \quad \forall a \in \mathcal{D}$ and satisfies the following conditions:
\begin{itemize}
    \item $P(x=a) \geq 0 \quad \forall a \in \mathcal{D}$.
    \item $\sum_{a \in \mathcal{D}} P(x=a) = 1$.
\end{itemize}
\subsection{Fréchet Inception Distance}
Fréchet Inception Distance (FID) \cite{FID} is a metric used for evaluating the quality of synthetic images generated by generative models such as GANs and AEs. It compares the distribution of generated images to that of a set of real-world images, with a score of 0 indicating a perfect match between the distribution of synthetic and real images. The FID metric has been employed in various works targeting biometric image generation, including IrisGAN~\cite{iris_gan}.

To compute the FID, first, an encoding of real and synthetic images is calculated using the Inception V3 network~\cite{inception_v3} without the classification layer. Subsequently, under the assumption that the encodings follow Gaussian distributions, mean values $\mu_r$ and $\mu_s$ as well as covariance matrices $\Sigma_r$ and $\Sigma_s$ are estimated for real and synthetic images, respectively. Finally, the FID is evaluated using the following formula:
\begin{equation*}
    FID = \left\| \mu_r - \mu_s \right\|_2^{2} + \text{tr} \left( \Sigma_r + \Sigma_s - 2 \left( \Sigma_r \Sigma_s \right)^{\frac{1}{2}} \right).
\end{equation*}

\subsection{Wasserstein Distance}
The Wasserstein distance \cite{OP-book-Gabriel} is a metric used for calculating the distance between probability distributions. It can be computed between two distributions, $p$ and $q$, as follows:
\begin{equation*}
    \mathcal{W}_\xi(p,q) = \left( \inf_{\gamma \in \mathcal{P}(p(x), q(x^\prime))} \mathbb{E}_{\gamma(x, x^\prime)} [d^\xi(x,x^\prime)]  \right)^\frac{1}{\xi}
\end{equation*}
where $d(x,x^\prime)$ represents the distance between $x$ and $x^\prime$, $\xi$ is a positive integer, and $\mathcal{P}(p(x), q(x^\prime))$ represents the joint distribution between $p$ and $q$.
Given that computing the infimum\footnote{The infimum of a subset $S$ of a partially ordered set $P$ is the greatest $p\in P$ that is less than or equal to each element of $S$, if such an element exists.} is often computationally challenging, entropy-regularized optimal transport was introduced in \cite{sinkhorn_distances} alongside the Sinkhorn algorithm. 

\section{Probability mass function sampling}
\label{contribution}
In this work, we introduce the Probability Mass Function Sampling (PMFS) method, an innovative approach for sampling from the latent space of \textsl{any} autoencoder model. This \textsl{post-training} method employs discrete density estimation through probability mass functions and quantization procedures. PMFS presents a solution to the challenge of continuous density estimation, which becomes complex due to the infinite nature of the space.

Formally, given a set of latent vectors \mbox{$\mathcal{Z} = \{z_i \in \mathbb{R}^d \}_{i=1}^n$} and a parameter $k$ indicating the number of uniform partitions or bins per dimension\footnote{The hyperparameter $k$ is chosen via a tuning process to minimize the FID metric (See Figure \ref{fig:evolution of fid}).}. Initially, we identify the maximum $max_j$ and minimum $min_j$ values per dimension $j\in\{1, \dots,d\}$. Subsequently, we compute the width of each partition $w_j=\frac{max_j-min_j}{k}$ for dimension $j$. The quantization step is then performed for each vector $z_i$ in the set $\mathcal{Z}$. This process assigns one of the $k^d$ global partitions in the space to each latent vector $z_i$. The quantized values $Q_{z_{i,j}}$ corresponding to the latent vector $z_{i}$ and each dimension $j$ are determined using the formula:
\begin{equation}
Q_{z_{i,j}} = \floor*{\frac{z_{i,j}-min_j}{w_j}} = \floor*{\frac{k(z_{i,j}-min_j)}{max_j-min_j}}.
\end{equation}

\noindent In Appendix \ref{example}, we provide an illustrative example of the proposed quantization step.

Lastly, we define our probability mass function as follows\footnote{This definition can be easily verified to represent a probability mass function.}:
\begin{equation}
    \label{parittion weight equation}
  P(x=p) = \frac{\text{\#vectors in the global partition $p$}}{n}
\end{equation}
where $p$ denotes a global partition. 
This formulation of the probability mass function can also be interpreted as a weight assigned to each global partition.
Hence, to sample a latent vector, a global partition $p$ is sampled according to its weight calculated using Equation~\ref{parittion weight equation}. Subsequently, given that each partition has upper and lower bounds $p_{min}$ and $p_{max}$, a vector $z$ is uniformly sampled from the partition, i.e., $z\sim \mathcal{U}_{[p_{min}, p_{max}]}(\cdot)$.
The proposed PMFS sampling method can be visualized as defining volumes with controllable dimensions around the known latent vectors, followed by uniform sampling from these volumes. PMFS sampling ensures that the generated samples can be accurately reconstructed into real-world data points, given their proximity to actual latent vectors. We provide a summary of the PMFS sampling model and its per-step time-complexity in Algorithm~\ref{alg:pmf}. Additionally, the related Python and Mathematica code is available in Appendix \ref{PMF sampling code}.

\begin{algorithm}[ht]
  \caption{Probability mass function sampling}
  \label{alg:pmf}
  \textbf{Input :} $\mathcal{Z} = \{z_i \in \mathbb{R}^d \}_{i=1}^n$ set of latent vectors, $k$ number of partition per dimension.
  
  \textbf{Output :} partitions and their weight;
  \begin{algorithmic}[1]
    \ForEach{$j \in \{1 \dots d \}$} \algorithmiccomment{\textcolor{ForestGreen}{$\mathcal{O}(n \times d)$}}
      \State $max_j$ = $\max_i \{z_{i,j}\}_{i=1}^n$;  \algorithmiccomment{\textcolor{ForestGreen}{$\mathcal{O}(n)$}}
      \State $min_j$ = $\min_i \{z_{i,j}\}_{i=1}^n$; \algorithmiccomment{\textcolor{ForestGreen}{$\mathcal{O}(n)$}}
    \EndForEach 

    \State $P = 0$ \algorithmiccomment{hashmap of partition weights.}
    
    \ForEach{$j \in \{1 \dots n \}$} \algorithmiccomment{\textcolor{ForestGreen}{$\mathcal{O}(n \times d)$}}
    
    \algorithmiccomment{Quantize vector $z_i$.}
      \State $Q_{z_i}$ = $Q(z_i, k, \{min_j\}_{j=1}^d, \{max_j\}_{j=1}^d)$  \algorithmiccomment{\textcolor{ForestGreen}{$\mathcal{O}(d)$}}

      \algorithmiccomment{Increment the weight of the partition $Q_{z_i}$.}
      \State $P[X=Q_{z_i}] = \frac{P[X=Q_{z_i}]  \times n + 1}{n} $ \algorithmiccomment{\textcolor{ForestGreen}{$\mathcal{O}(1)$}}
    \EndForEach 
  \end{algorithmic}
\end{algorithm}

Analyzing and comparing the PMFS sampling algorithm to the Gaussian Mixture Model (GMM) sampling algorithm, we observe that in terms of time complexity, we transition from $\mathcal{O}(n \times d \times k \times i)$ to $\mathcal{O}(n \times d)$, which constitutes a significant improvement. Moreover, we shift from relying on the infinite Expectation-Maximization (EM) algorithm~\cite{EM} with the number of iterations $i$ as a stopping criterion to a finite algorithm that depends solely on the number of latent vectors $n$ and their dimension $d$.

Furthermore, for performance optimization, we retain only bins containing data points by utilizing a hashmap to store the weights. This is a strategy to avoid an exponential growth in the number of partitions as the latent dimension $d$ increases, given the $k^d$ global partitions generated. Consequently, we only maintain in memory the weights of existing partitions. This approach is equivalent to preserving all bins without sacrificing performance nor generality, as global partitions lacking data samples possess a weight of zero and therefore never get selected.

To summarize, the PMFS method facilitates density estimation and the sampling of high-probability vectors. When these vectors are reconstructed by the decoder network, they yield synthetic data points that closely resemble real-world data (See to Section \ref{exp} and Appendix \ref{synthetic images}). Also, the estimation process in our model operates with a time complexity of $\mathcal{O}(n \times d)$. This stands as a significant improvement over Gaussian mixture model sampling, which entails a time complexity of $\mathcal{O}(n \times d \times k \times i)$. Finally, we address the challenge of an exponentially growing number of partitions by utilizing a hash map to retain only pertinent partitions.

\section{Experiments}
\label{exp}
This section presents experimental results validating the proposed PMFS sampling method described in Section~\ref{contribution}, compared to GMM sampling. In Section \ref{overall-archi} we present an overview of the autoencoder model architectures employed in this work. Appendix \ref{exp-setup} and \ref{datasets} present our experimental setup and a description of the used datasets. Moreover, in Appendix \ref{image reconstruction}, we provide samples of image reconstructions using different autoencoder models. Furthermore, we lay out a qualitative comparison on the impact of the sampling method on the synthetic images in Appendix~\ref{synthetic images}.

\subsection{Model architectures}
\label{overall-archi}
The autoencoder models used is this work contain an encoder network with convolutional layers and a linear layer which allows operating on images. The decoder network comprises a non-linear layer and a set of transpose convolutional layers to reconstruct the input images (See Figure~\ref{fig:archi}, ample network architecture details are provided in Appendix~\ref{archis}).
\begin{figure}[H]
    \centering
    \includegraphics[width=\columnwidth]{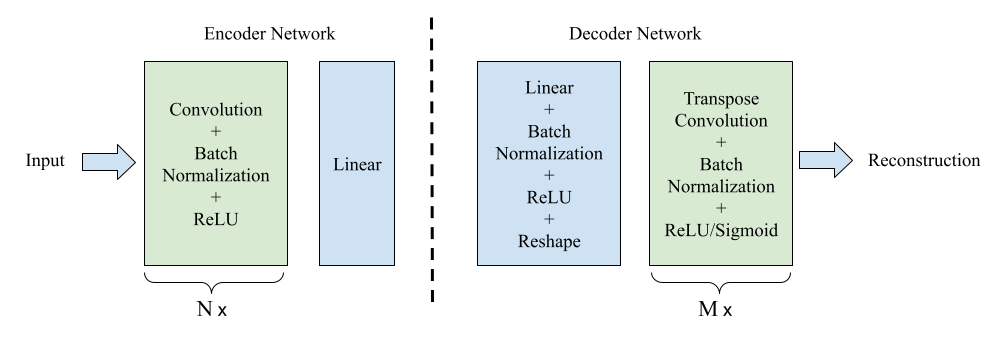}
    \caption{Overview of the models' architecture used in this work.}
    \label{fig:archi}
\end{figure}

\subsection{Training parameters}
\label{train-params}
The MOBIUS \cite{rot2018deep,rot2020deep,vitek2023exploring, ssbc2020, vitek2020comprehensive}, MNIST \cite{mnist}, and CelebA \cite{celeba} datasets used in this work were maintained at their original resolutions. The different models were trained for $50$ epochs using the Adam optimizer with a learning rate of $10^{-3}$ and $\beta_1=0.9, \beta_2=0.999$, and a batch size of $64$. However, for the CelebA dataset, we trained our models for $100$ epochs with a learning rate of $10^{-4}$. The rest of the parameters remained unchanged. We ensured the convergence of all models. In terms of latent dimensionality, we employed sizes of $32$, $256$, and $256$ for the MNIST, CelebA, and MOBIUS datasets, respectively, for all models. Moreover, we used the latent vectors of the validation set images to fit the GMM and PMFS models.
The models utilized in this work encompass the Vanilla Autoencoder \cite{Goodfellow-et-al-2016}, Variational Autoencoder \cite{VAE}, $\beta$-VAE~\cite{beta-vae} with $\beta=2$, Wasserstein Autoencoder \cite{wae} with $\beta=100$, and InfoVAE~\cite{infovae} with $\alpha=0, \lambda=1000$. The hyperparameters of the networks are those yielding the best results in the original papers introducing each model, except for $\beta$-VAE, where we set $\beta=2$ to better minimize the reconstruction~loss.

\subsection{Results and discussion}
\label{results}
\begin{figure*}
    \begin{subfigure} {0.33\textwidth}
         \centering
         \includegraphics[width=\columnwidth]{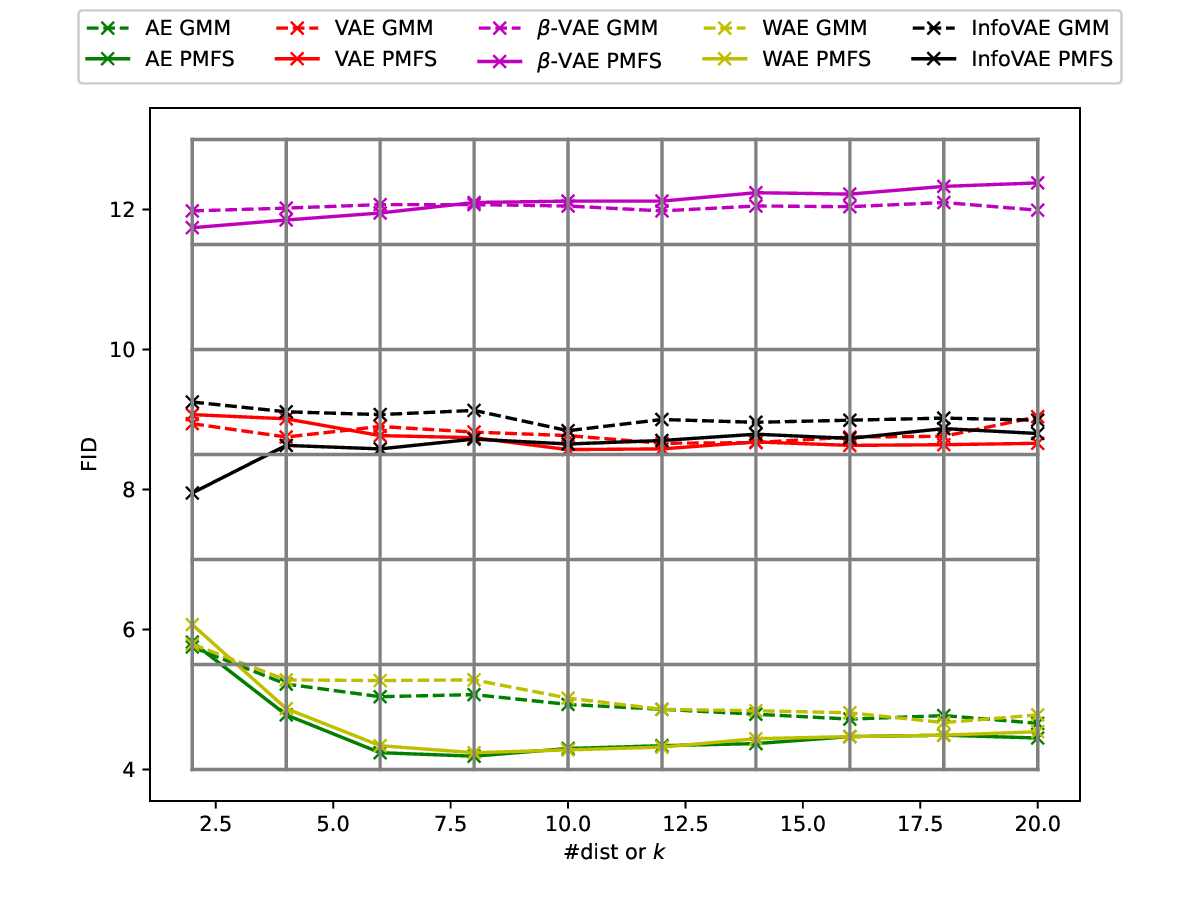}
         \caption{MNIST data set.}
     \end{subfigure}
     \begin{subfigure} {0.33\textwidth}
         \centering
         \includegraphics[width=\columnwidth]{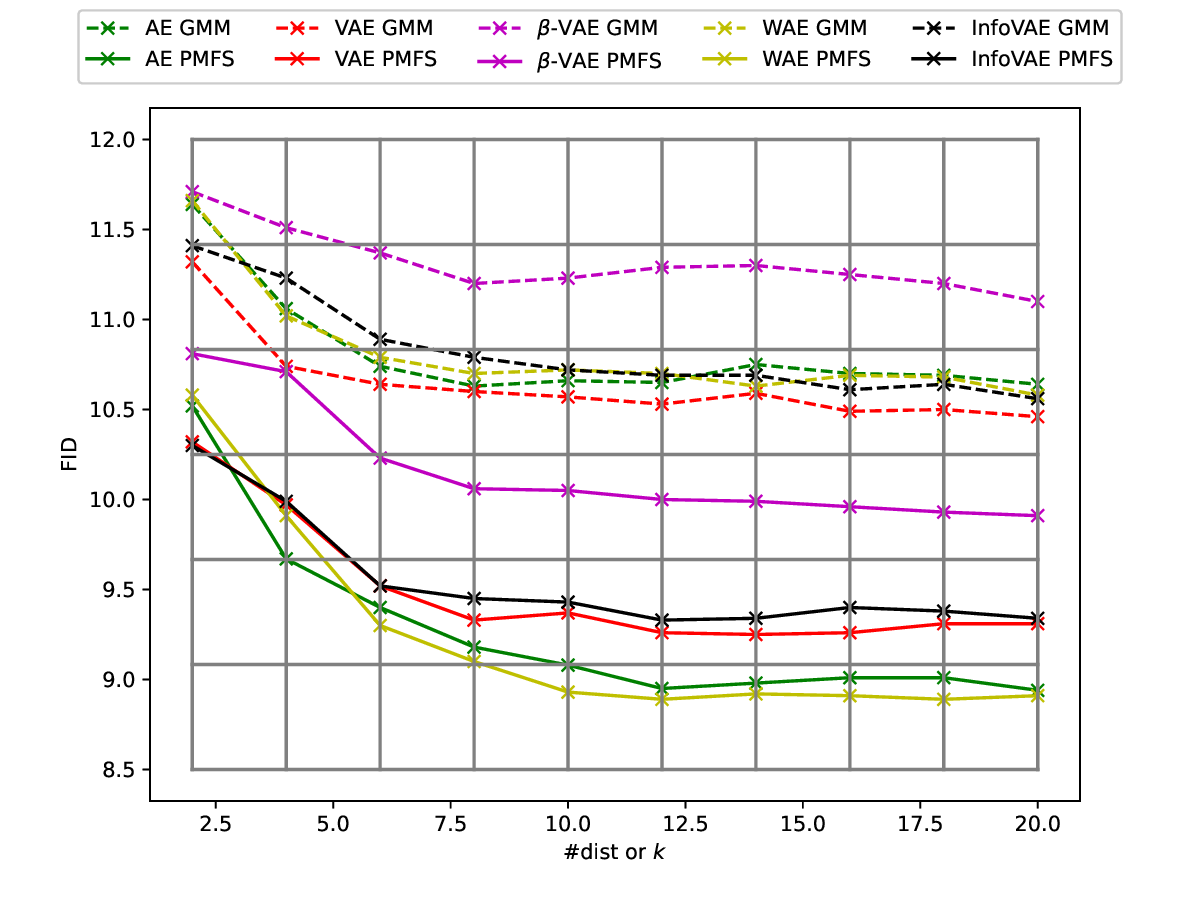}
         \caption{CelebA data set.}
     \end{subfigure}
    \begin{subfigure} {0.33\textwidth}
         \centering
         \includegraphics[width=\columnwidth]{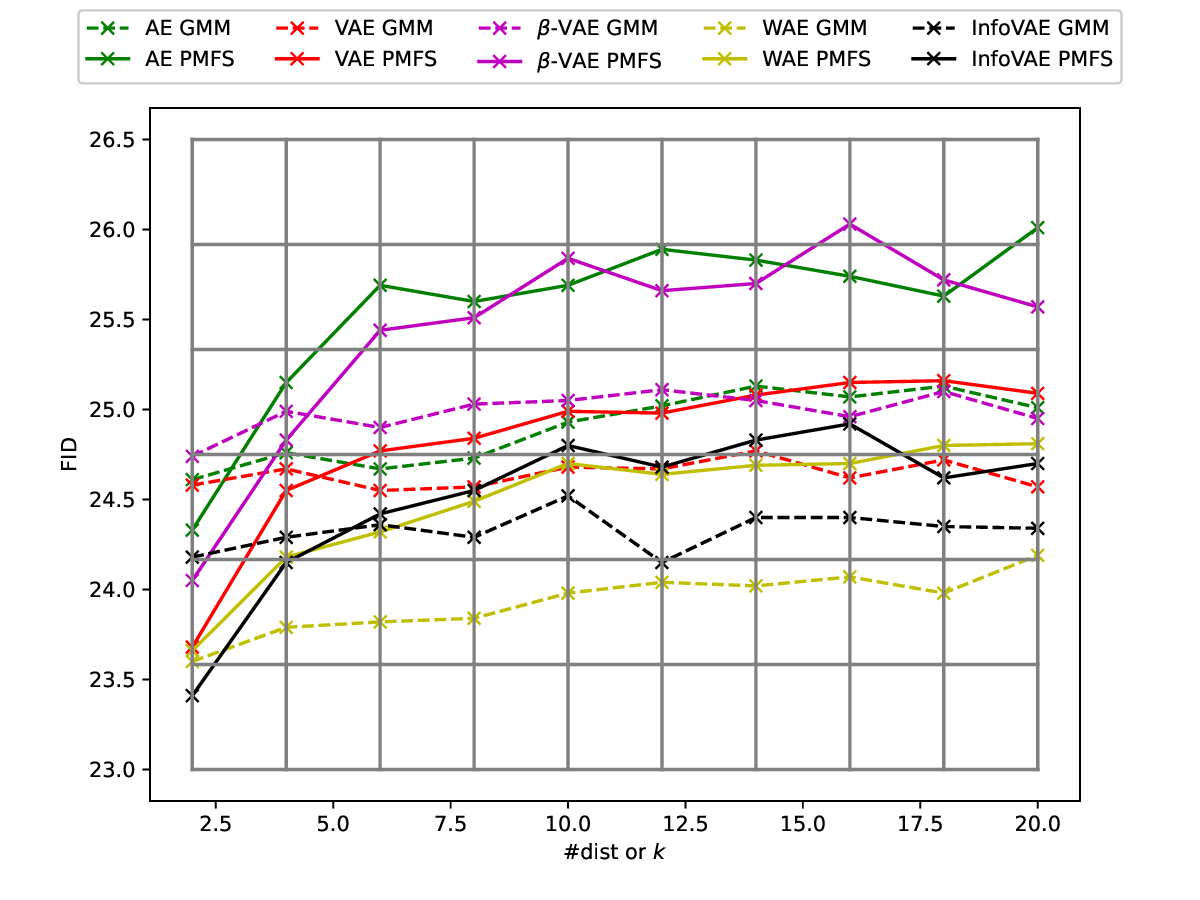}
         \caption{MOBIUS data set.}
     \end{subfigure}
     \caption{The evolution of the FID metric in relation to the number of partitions in PMFS or the number of distributions in a GMM. Dashed and solid lines respectively illustrate the FID evolution using GMM and PMFS sampling methods. The colors \textcolor{ForestGreen}{green}, \textcolor{red}{red}, \textcolor{magenta}{magenta}, \textcolor{lime}{yellow}, and black correspond to the Vanilla Autoencoder, Variational Autoencoder (VAE), $\beta$-VAE, Wasserstein Autoencoder (WAE), and InfoVAE models, respectively. This figure is best viewed in color.
     }
     \label{fig:evolution of fid}
\end{figure*}

\begin{figure*}
    \centering
    \begin{subfigure} {0.33\textwidth}
         \centering
         \includegraphics[width=\columnwidth]{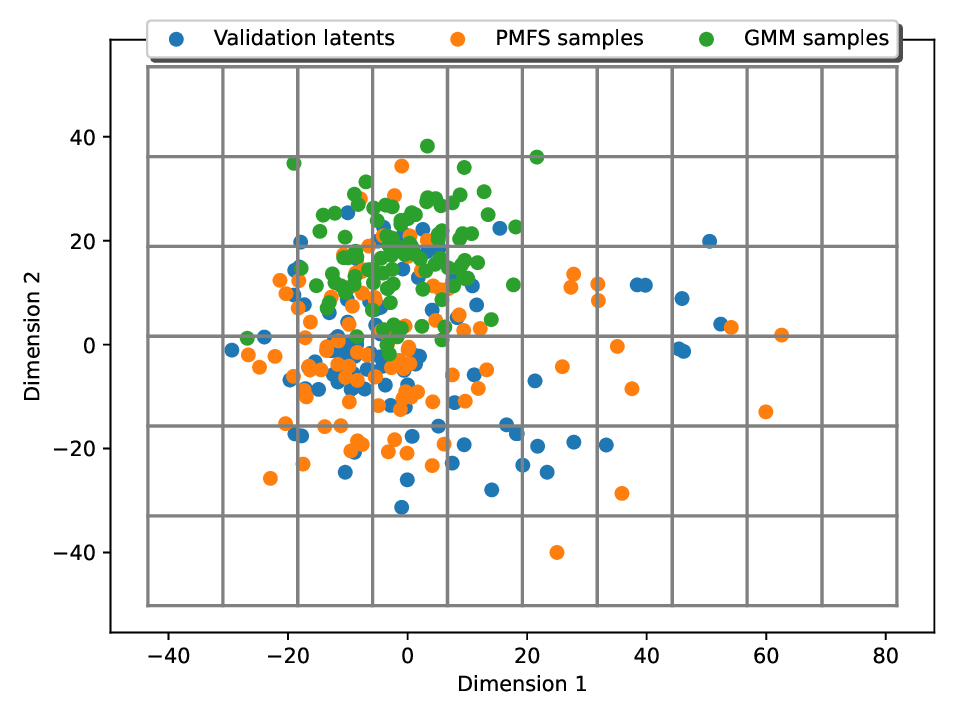}
         \caption{Latent space distribution on MNIST.}
     \end{subfigure}
     \hfill
     \begin{subfigure} {0.33\textwidth}
         \centering
         \includegraphics[width=\columnwidth]{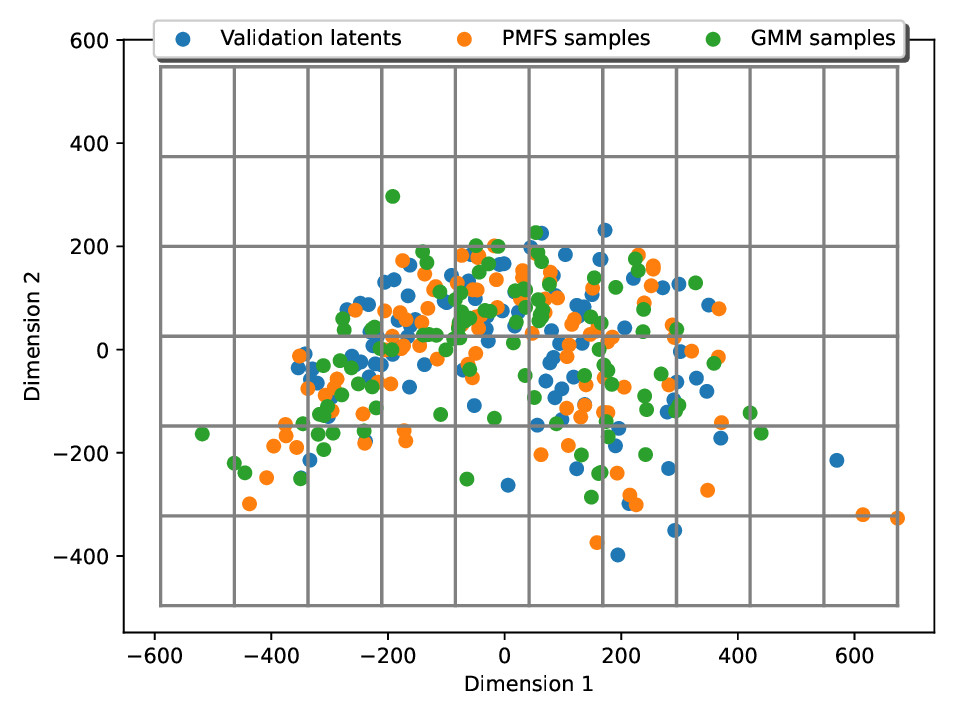}
         \caption{Latent space distribution on CelebA.}
     \end{subfigure}
     \hfill
     \begin{subfigure} {0.33\textwidth}
         \centering
         \includegraphics[width=\columnwidth]{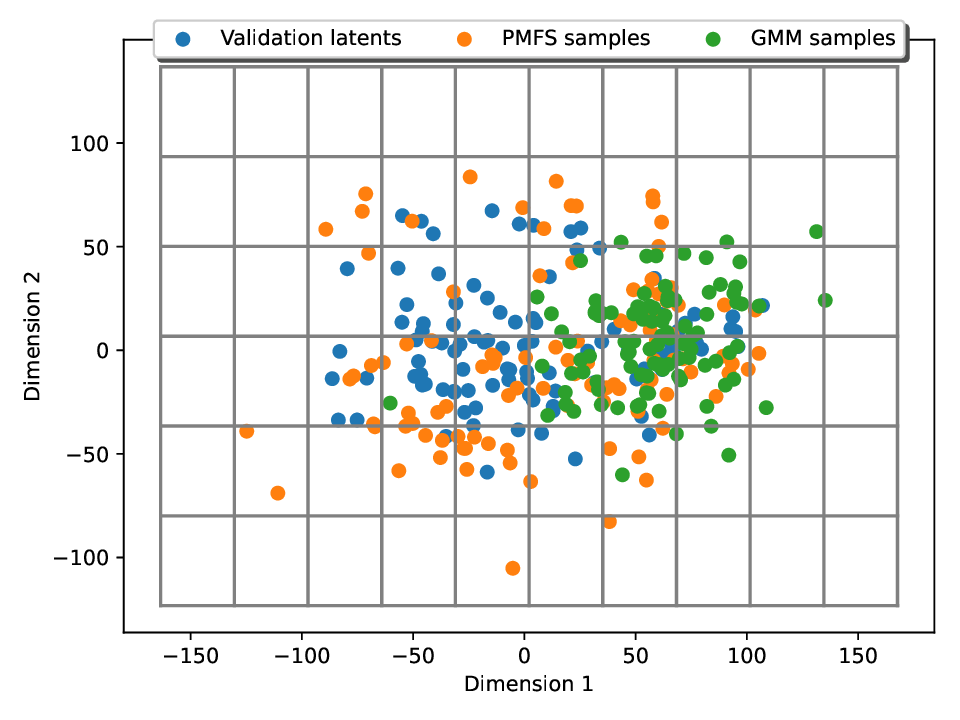}
         \caption{\mbox{Latent space distribution on MOBIUS.}}
     \end{subfigure}
     \caption{An illustration of the distribution of the latent vectors from the validation set of the Vanilla Autoencoder, along with samples generated using GMM and PMFS sampling. These figures portray a dimensionality reduction through PCA into the two-dimensional space \(\mathbb{R}^{2}\), encompassing the first hundred images from the validation set as well as a hundred samples extracted from the GMM and PMFS sampling methods. This figure is best viewed in color.
     }
     \vspace{-.3cm}
     \label{fig:distributions}
\end{figure*}

Following the training of various autoencoder-based generative models, notably the Vanilla Autoencoder \cite{Goodfellow-et-al-2016}, Variational Autoencoder \cite{VAE}, $\beta$-VAE \cite{beta-vae}, Wasserstein Autoencoder~\cite{wae}, and InfoVAE \cite{infovae} on the MNIST \cite{mnist}, CelebA~\cite{celeba}, and MOBIUS \cite{rot2018deep,rot2020deep,vitek2023exploring, ssbc2020, vitek2020comprehensive} datasets, we computed the Fréchet Inception Distance FID metric. The obtained results are presented in Table \ref{tab:fid_results}. Furthermore, Figure \ref{fig:evolution of fid} illustrates the progression of the FID metric in relation to the number of distributions for GMM sampling and the number of partitions for PMFS sampling. It's important to note that for calculating the FID metric, we generated a quantity of synthetic images equivalent to the number of images in the respective test set for each dataset. Lastly, Figure \ref{fig:distributions} displays the distribution of the latent space of the validation set, providing a comparison with samples drawn from GMM and PMFS models.

\begin{table}[ht]
    \centering
    \caption{A comparison between GMM sampling and PMFS based on the lowest FID, Wasserstein distance (denoted by $\mathcal{W}$), and model fitting time in seconds (denoted by \(T\)). The comparison is performed across the MNIST, CelebA, and MOBIUS datasets. The columns labeled \#dist and \(k\) respectively signify the number of distributions utilized in GMM sampling and the number of partitions per dimension in the PMFS sampling method, aiming to minimize the FID metric. In the table, the most favorable values are highlighted in \textbf{bold}.
    }
    
    \resizebox{\columnwidth}{!}{%
    \begin{tabular}{|c|c|cccc|cccc|}
    \toprule
         & \multirow{2}{*}{\textbf{Model}} & \multicolumn{4}{c|}{\textbf{GMM}} & \multicolumn{4}{c|}{\textbf{PMFS}}  \\
         &  & FID $\downarrow$ & $\mathcal{W} \downarrow$ & T $\downarrow$ & \#dists & FID $\downarrow$ & $\mathcal{W} \downarrow$ & T $\downarrow$ & $k$  \\
        \midrule
        \multirow{5}{*}{\rotatebox[origin=c]{90}{MNIST}} & AE & \phantom{0}4.66 & 839.77 & 25.64 & 20 & \textbf{\phantom{0}4.19} & \textbf{551.81} & \textbf{0.02} & \phantom{0}8 \\
        & VAE & \phantom{0}8.66  & \textbf{13.63} & 20.59 & 12 & \textbf{\phantom{0}8.57} & 14.16 & \textbf{0.07} & 10\\
        & $\beta$-VAE & 11.98 & \textbf{13.50} & 24.64 & 12 & \textbf{11.74} & 53.71 & \textbf{0.06} & \phantom{0}2\\
        & WAE & \phantom{0}4.67 & 880.54 & 29.24 & 18 & \textbf{\phantom{0}4.24} & \textbf{570.55} & \textbf{0.04} & \phantom{0}8\\
        & InfoVAE & \phantom{0}8.84 & \textbf{13.03} & 18.81 & 10 & \textbf{\phantom{0}7.95} & 49.18 & \textbf{0.08} & \phantom{0}2\\\midrule
        
         \multirow{5}{*}{\rotatebox[origin=c]{90}{CelebA}} & AE & 10.63 & $4.6 \times 10^{5}$ & 95.56 & \phantom{0}8 & \textbf{\phantom{0}8.94} & $\mathbf{3.3 \times 10^{5}}$ & \textbf{0.11} & 20 \\
        & VAE & 10.49 & 190.93 & 65.31 & 16 & \textbf{\phantom{0}9.25} & \textbf{159.14} & \textbf{0.18} & 14 \\
        & $\beta$-VAE & 11.10 & 196.48 & 50.01 & 20 & \textbf{\phantom{0}9.91} & \textbf{170.28} & \textbf{0.16} & 20 \\
        & WAE & 10.58 & $4.9 \times 10^{5}$ & 52.86 & 20 & \textbf{\phantom{0}8.89} & $\mathbf{3.8 \times 10^{5}}$ & \textbf{0.15} & 12 \\
        & InfoVAE & 10.56 & 193.85 & 48.90 & 20 & \textbf{\phantom{0}9.33} & \textbf{159.26} & \textbf{0.18} & 12 \\\midrule
        
        \multirow{5}{*}{\rotatebox[origin=c]{90}{MOBIUS}} & AE & 24.61 & $\mathbf{6.4 \times 10^{3}}$ & \phantom{0}1.59 & \phantom{0}2 & \textbf{24.33} & $2 \times 10^{4}$ & \textbf{0.02} & \phantom{0}2 \\
        & VAE & 24.55 & \textbf{528.51} & \phantom{0}1.48 & \phantom{0}6 & \textbf{23.68} & 1765.73 & \textbf{0.06} & \phantom{0}2\\
        & $\beta$-VAE & 24.74 & \textbf{303.68} & \phantom{0}1.69 & \phantom{0}2 & \textbf{24.05} & 836.28 & \textbf{0.02} & \phantom{0}2\\
        & WAE & \textbf{23.60} & $\mathbf{5.9 \times 10^{3}}$ & \phantom{0}1.57 & \phantom{0}2 & 23.66 & $1.8 \times 10^{4}$ & \textbf{0.03} & \phantom{0}2\\
        & InfoVAE & 24.18 & \textbf{387.27} & \phantom{0}1.78 & \phantom{0}2 & \textbf{23.41} & 1156.14 & \textbf{0.05} & \phantom{0}2 \\\bottomrule 
    \end{tabular}
    }
    \label{tab:fid_results}
    \vspace{-.3cm}
\end{table}

\begin{figure*}[tb]
    \begin{subfigure} {0.49\textwidth}
         \centering
         \includegraphics[width=.95\textwidth]{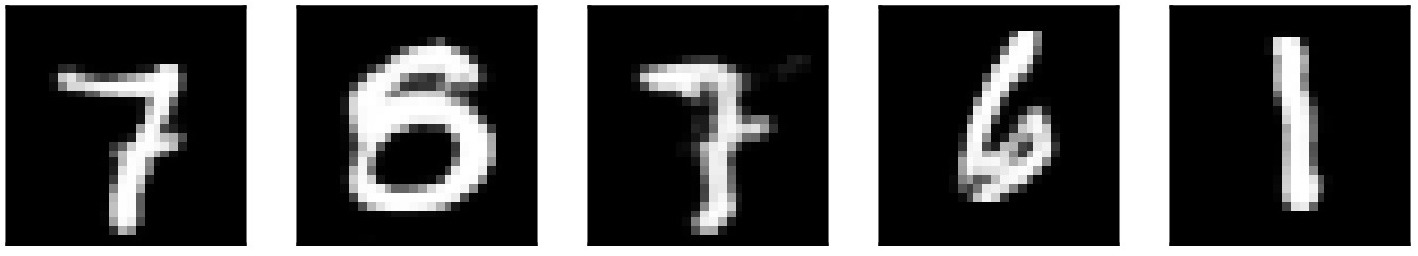}
         \caption{MNIST GMM samples.}
     \end{subfigure}
     \hfill
     \begin{subfigure} {0.49\textwidth}
         \centering
         \includegraphics[width=.95\textwidth]{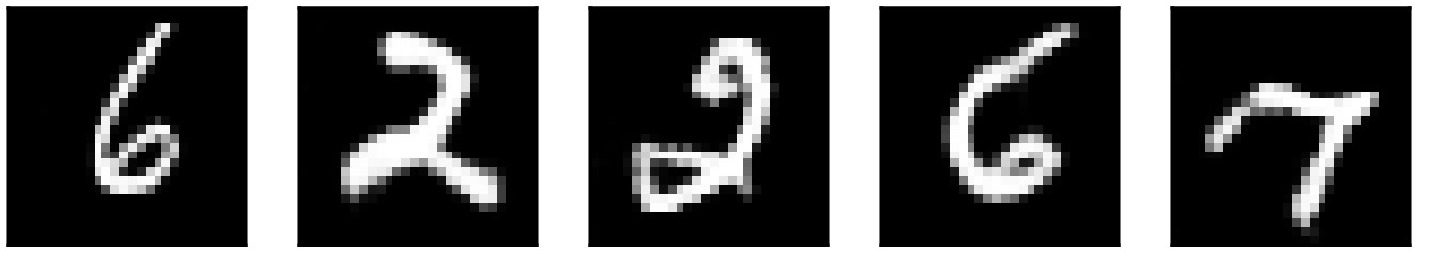}
         \caption{MNIST PMFS samples.}
     \end{subfigure}
     \begin{subfigure} {0.49\textwidth}
         \centering
         \includegraphics[width=.95\textwidth]{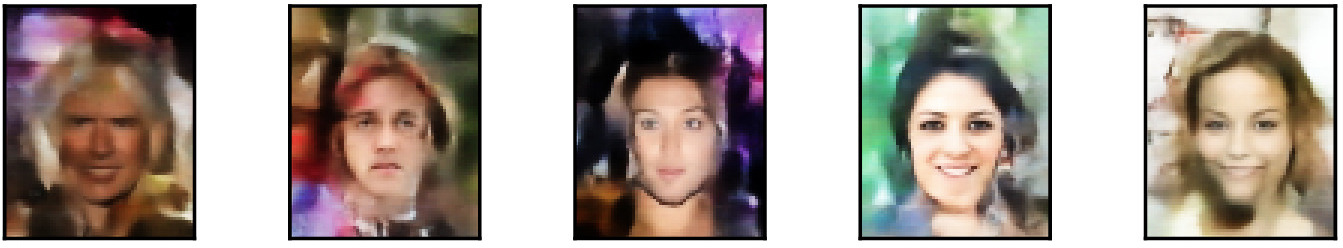}
         \caption{CelebA GMM samples.}
     \end{subfigure}
     \hfill
     \begin{subfigure} {0.49\textwidth}
         \centering
         \includegraphics[width=.95\textwidth]{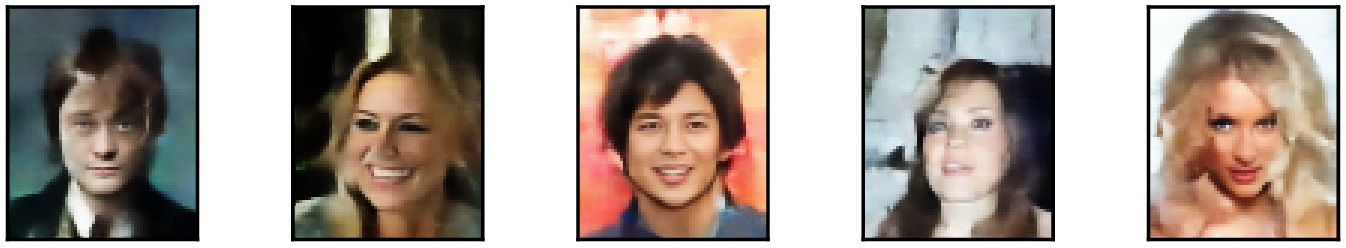}
         \caption{CelebA PMFS samples.}
     \end{subfigure}

     \begin{subfigure} {0.49\textwidth}
         \centering
         \includegraphics[width=.95\textwidth]{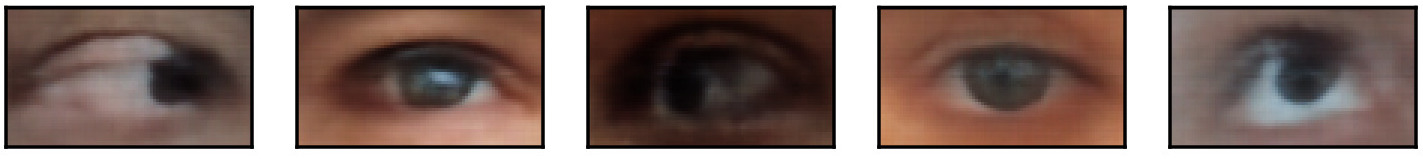}
         \caption{MOBIUS GMM samples.}
     \end{subfigure}
     \hfill
     \begin{subfigure} {0.49\textwidth}
         \centering
         \includegraphics[width=.95\textwidth]{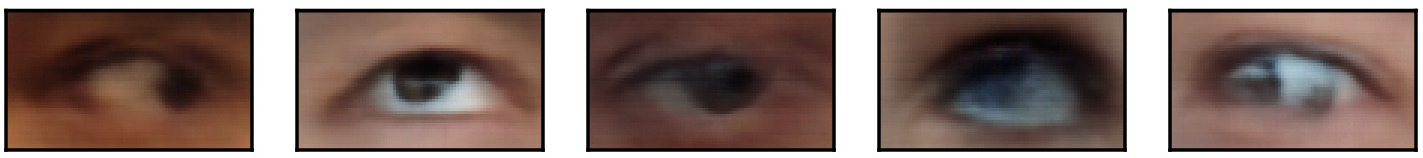}
         \caption{MOBIUS PMFS samples.}
     \end{subfigure}
     
     \caption{Sample synthetic images from a Vanilla Autoencoder model on the MNIST, CelebA and MOBIUS data sets using GMM and PMFS sampling. Figure better viewed in color.}
     \vspace{-.3cm}
     \label{fig:samples synthetic images}
\end{figure*}

From Table \ref{tab:fid_results} and Figure \ref{fig:evolution of fid}, it becomes evident that, in terms of GMM sampling, the optimal choice for the number of distributions (\#dist) is rarely the same as the number of classes in the dataset, even for relatively simple datasets like MNIST. As a result, relying on the heuristic of setting the number of classes as the number of distributions in a GMM, as is commonly done \cite{gmm-10}, is likely to yield suboptimal outcomes. Furthermore, a comparison of the lowest FID values achieved by GMM sampling with those obtained through PMFS sampling underscores the enhancements introduced by PMFS sampling across all models and datasets. This improvement can be attributed to the strategy of defining a neighborhood around each latent vector and sampling from within this neighborhood, leading to the acquisition of superior, high-probability samples that can be effectively reconstructed by the decoder network into realistic data points. This contrasts with GMM sampling, which operates on distributions encompassing the entire space and might yield samples lying outside the regions reconstructable by the decoder network. Such a characteristic is crucial in biometric image synthesis, where the goal is to generate synthetic images that closely resemble real-world counterparts.

Furthermore, Table \ref{tab:fid_results} reveals substantial gains in terms of the FID metric when applying our proposed PMFS sampling method to deterministic models, particularly the Vanilla Autoencoder \cite{Goodfellow-et-al-2016} and the Wasserstein Autoencoder~\cite{wae}. This trend can be attributed to the stochastic learning phase in these models, which capitalizes on the vicinity of each mean latent vector corresponding to an image. Consequently, our method, rooted in the concept of neighborhoods, yields slightly reduced efficacy for this model type. However, it still outperforms the GMM sampling method in terms of FID. Consequently, when employing the PMFS sampling method on biometric data, the outcome is a collection of synthetic images resembling more closely the actual instances. Additionally, a comparison of the model fitting time between PMFS and GMM sampling reveals a substantial improvement, quantifiable in terms of orders of magnitude, thus corroborating the anticipated theoretical disparity in time complexity.

Moreover, we conducted an investigation into the distribution of drawn samples via GMM sampling and PMFS, comparing them to the underlying distribution of latent vectors generated by the validation dataset. This investigation was executed quantitatively using the Wasserstein distance \cite{OP-book-Gabriel} (see the $\mathcal{W}$ column in Table \ref{tab:fid_results}) and qualitatively (see Figure \ref{fig:distributions}). From the $\mathcal{W}$ column in Table \ref{tab:fid_results}, it is evident that, particularly for large datasets like MNIST and CelebA, PMFS has the capacity to generate distributions which closely approximate the real ones, demonstrating an improvement over GMM sampling. This holds true, especially when the value of \(k\) is high. Consequently, for smaller datasets like MOBIUS, GMM sampling exhibits superior performance over PMFS due to the low number of partitions \(k\) employed in PMFS, aiming to maximize FID while compensating for the distribution discrepancy between the training and testing sets. These findings are visually supported by Figure~\ref{fig:distributions} and Appendix \ref{dists}.

Upon analyzing Figure \ref{fig:evolution of fid}, it becomes evident that, for all the datasets, there exists a specific number of partitions \(k\) for PMFS sampling that yields the lowest FID outcome, irrespective of the number of distributions employed in GMM sampling. Particularly on the CelebA dataset, known for face generation, a substantial advantage of PMFS sampling over GMM sampling is apparent. Here, a notable margin separates the images generated using GMM sampling from those produced via PMFS sampling. Additionally, it's noticeable that the optimal value for the number of partitions~\(k\) in PMFS sampling varies depending on the dataset.

Figure \ref{fig:samples synthetic images} showcases samples of synthetic images generated through GMM and PMFS sampling using a Vanilla Autoencoder model (Additional images are provided in Appendix \ref{synthetic images}). From this figure, we notice that the images generated with PMFS sampling possess a heightened realism. Notably, when PMFS sampling is applied to the CelebA dataset. For example, the first image depicts the actor \textsl{Daniel Radcliffe (Harry Potter)} without glasses. This observation underscores how PMFS sampling introduces data augmentation and modifications to images present in the training dataset. This enhancement emphasises the significance of our method in biometric image synthesis and data augmentation on biometric datasets. 

Furthermore, in Table \ref{FIQA}, we present a face image quality assessment on the synthetic images generated using PMFS and GMM sampling via the SDD-FIQA metric \cite{sdd_fiqa}. Since the SDD-FIQA metric uses a model which was not trained on the CelebA data set \cite{celeba}, we start by establishing a baseline by calculating the value of SDD-FIQA metric on the original untouched images of the CelebA data set. Then we generate $10,000$ synthetic images per sampling strategy, notably GMM sampling and PMFS, and per autoencoder model. The obtained results are presented in Table~\ref{FIQA}. From this table we observe improvements in terms of synthetic face image quality on all models when samples are generated through PMFS which emphasizes the performance gains of PMFS over GMM sampling.
\begin{table}[H]
    \centering
    \caption{Face image quality assessment on the CelebA data set. In the table, the most favorable values are highlighted in \textbf{bold}.}
    \resizebox{0.7\columnwidth}{!}{%
    \begin{tabular}{|c|cc|}
        \toprule
        \multirow{2}{4cm}{\centering Maximum obtainable value on CelebA data set} & \multicolumn{2}{|c|}{\multirow{2}{*}{41.55}} \\
        &&\\\midrule
        \textbf{Model} & \textbf{GMM} & \textbf{PMFS} \\\midrule
        Vanilla Autoencoder & 39.28 & \textbf{39.85} \\
        VAE & 38.78 & \textbf{39.43} \\
        $\beta$-VAE & 38.71 & \textbf{38.79} \\
        WAE & 37.60 & \textbf{38.85} \\
        InfoVAE & 38.68 & \textbf{39.39} \\
        \bottomrule
    \end{tabular}
    }\vspace{-.3cm}
    \label{FIQA}
\end{table}

In this section, we have presented a quantitative comparison through the FID metric and a qualitative comparison via synthetic images between the GMM sampling method and our proposed method. These comparisons were conducted across five distinct autoencoder-based generative models and three diverse datasets, including two biometric datasets. Throughout these experiments, PMFS sampling consistently outperformed GMM sampling, showcasing the lowest FID values, achieving fitting times several orders of magnitude faster, producing more lifelike synthetic images across all scenarios, and on large datasets with a high number of partitions generating samples with distributions that closely resemble real-world ones.

\section{Conclusion}
\label{conclusion}
In this study, our focus revolves around the task of sampling from the latent space of autoencoder-based generative models. Our primary objective is to extract samples from regions characterized by high density, ensuring that these samples can be subsequently reconstructed into images that possess a realistic quality. To accomplish this goal, we introduce a novel sampling algorithm that leverages the concept of probability mass functions coupled with a quantization process. Notably, this algorithm boasts a time complexity of $\mathcal{O}(n\times d)$, resulting in a substantial acceleration compared to Gaussian mixture model (GMM) sampling.
The core principle underlying our algorithm entails the definition of neighborhoods surrounding each latent encoding corresponding to input data points. Subsequently, the sampling process targets these neighborhoods, guaranteeing that the sampled latent vectors fall within regions of elevated probability. Consequently, these vectors can be seamlessly transformed into genuine images. Our experimentation covers a diverse array of scenarios, spanning three distinct datasets—MNIST for digit generation, CelebA for face generation, and MOBIUS for ocular generation—as well as five autoencoder-based generative models. The empirical results across our experiments consistently showcase the superiority of the PMFS sampling method in comparison to Gaussian mixture model (GMM) sampling. Impressively, this superiority is especially pronounced on the biometric datasets—CelebA and MOBIUS—where PMFS exhibits a remarkable enhancement of $1.69$ and $0.87$ in terms of the Fréchet inception distance (FID), respectively, when compared to GMM sampling. These findings underscore the exceptional efficacy of our novel sampling algorithm, particularly in generating high-quality images in diverse contexts, with a marked emphasis on biometric images. Moreover, when considering the Wasserstein distance, PMFS outperforms GMMs in approximating the true latent distribution.
Prospective extensions of our work could encompass enhancements aimed at improving the image quality produced by autoencoder models by mitigating the blur introduced by the $\ell_2$ norm \cite{Goodfellow-et-al-2016}. Additionally, we aspire to differentiate between whether the generated synthetic images serve as data augmentations over the training set or constitute an entirely new dataset, thereby necessitating the quantification of data leakage. This perspective is particularly salient given the sensitive nature of biometric data, which warrants stringent privacy considerations. In summary, these insights lay the groundwork for further exploration and refinement of the PMFS method.

\section*{Acknowledgement}
The authors extend their gratitude to Stéphane Ayache, Laurent Galathée, and Adnan Ben Mansour for their invaluable input and constructive feedback on the content of this paper.
Furthermore, the authors acknowledge the generous access granted to them for the utilization of High-Performance Computing (HPC) resources available through MesoPSL. This valuable resource is funded by the Région Île-de-France and is a part of the Equip@Meso project (reference \mbox{ANR-10-EQPX-29-01}). The Equip@Meso project is a constituent of the \textsl{programme investissements d’avenir}, which is overseen by France's \textsl{Agence nationale pour la recherche}.
The MOBIUS Data set images employed in this research have been generously provided by the University of Ljubljana, Slovenia~\cite{rot2018deep,rot2020deep,vitek2023exploring, ssbc2020, vitek2020comprehensive}.

{\small
\bibliographystyle{ieee}

}

\newpage
\appendix
\onecolumn
\begin{center}
	{\LARGE \textbf{Supplementary material}}
\end{center}

\section{Quantization example}
\label{example}
Given the dimension $d=3$ latent vector $z_i=[1.5; 2.6; 8]$ and the two vectors $[-19; -5; 0]$, $[5.7; 3; 20]$ representing the minimums and maximums for each dimension, respectively, and $k=10$. The quantized components of the vector $Q_{z_i}$ corresponding to the latent  vector $z_i$ are calculated as follows : 
\begin{align*}
    Q_{z_{i,1}} &= \floor*{\frac{10(1.5 + 19)}{5.7+19}} = \floor*{\frac{10\times 20.5}{24.7}} = \floor*{8.29} = \mathbf{8} \\
    Q_{z_{i,2}} &= \floor*{\frac{10(2.6+5)}{3+5}} = \floor*{\frac{10\times 7.6}{8}} = \floor*{9.5} = \mathbf{9} \\ 
    Q_{z_{i,3}} &= \floor*{\frac{10(8-0)}{20-0}} = \floor*{\frac{10\times 8}{20}} = \floor*{4} = \mathbf{4}.
\end{align*}

\noindent Thus, the quantized vector $Q_{z_i}$ is $[8; 9; 4]$. This means the vector $z_i$ belongs to the global partition $[8; 9; 4]$ (partition $8$ in the first dimension, partition $9$ in the second dimension and partition $4$ in the last dimension). We point out that we start the numbering of partitions from zero.

\section{Probability mass function sampling code}
\label{PMF sampling code}
\noindent Python code:
\begin{lstlisting}language={python}
def PMFS_sampling(z, k):
    max_, min_ = np.max(z, axis=0, keepdims=True), np.min(z, axis=0, keepdims=True)
    partition_sizes = (max_ - min_)/k
    z = ((z-min_)/partition_sizes).astype(np.int8)
    unique, counts = np.unique(z, axis=0,return_counts=True)
    bin_edges = (min_, max_, num=k)
    return unique, counts, bin_edges
\end{lstlisting}

\usemintedstyle{mathematica}
\noindent Mathematica code:
\begin{lstlisting}[extendedchars=true,language=Mathematica]
MFSSampling[z_List, k_Integer] := Module[{max, min, partitionSizes, quantizedZ, unique, counts, 
   binEdges}, max = Max[z]; min = Min[z];
  partitionSizes = (max - min)/k;
  quantizedZ = Floor[(z - min)/partitionSizes];
  {unique, counts} = Transpose[Tally[quantizedZ]];
  binEdges = {min, max, k};
  {unique, counts, binEdges}]
\end{lstlisting}

\section{Experimental setup}
\label{exp-setup}
To train the models for this work, we used the Pytorch \cite{pytorch} and Pytorch Lightning \cite{pytorch_lightning} libraries in conjunction with Torchvision \cite{torchvision} to obtain the training benchmark data sets. We conducted experiments on Nvidia Tesla V100 GPUs with $32$ GB of VRAM, available in the MesoPSL computing cluster. With these GPUs, training on the MNIST, CelebA and MOBIUS data sets took approximately $1$ minute, $14$ minutes and $25$ minutes per epoch respectively, while computing the FID required approximately $30$ minutes per data set. 

The Numpy \cite{numpy} and the Scikit-learn \cite{scikit-learn} libraries are respectively used to implement the PMFS and GMM sampling strategies. The Matplotlib library \cite{matplotlib} is used to generate the figures in this work.


\section{Datasets}
\label{datasets}
\subsection{MNIST}
The MNIST dataset \cite{mnist}, comprises ten classes of grayscale images, each with a size of $28\times 28$, which we employ for training and testing our models. The dataset is divided into~: a training set with $50,000$ images, a validation set with $10,000$ images, and a test set with $10,000$ images. This dataset serves as a standard benchmark for computer vision tasks, particularly for evaluating models on simple, low-resolution grayscale images.
\vspace{-.3cm}

\subsection{CelebA}
The Celebrity Faces dataset \cite{celeba}, encompasses colored images of celebrity faces with a size of $178\times 218$, which we employ for training and testing our models. It comprises a total of $200,000$ images. In our experiments, we utilize the original train, validation, and test dataset split provided with the dataset. This dataset serves as a standard benchmark for computer vision applications, particularly in the domain of face generation, rendering it highly relevant to the field of face biometrics.

\subsection{MOBIUS}
The MOBIUS dataset \cite{rot2018deep,rot2020deep,vitek2023exploring, ssbc2020, vitek2020comprehensive} is a compilation of $16,717$ colored ocular images with dimensions \mbox{$1,700\times 3,000$}, which we utilize for training and testing our models. During our experiments, we removed blurry images, resulting in a total of $14,331$ images. We partitioned the dataset into training, validation, and test subsets, consisting of $9,172$, $2,293$, and $2,866$ images, respectively. The images from the MOBIUS dataset employed in this work have been generously provided by the University of Ljubljana, Slovenia, representing a typical biometric dataset suitable for image generation tasks.

\section{Model architecture details}
\label{archis}
The models we trained in this work have the architectures details in Tables \ref{tab:archi MNIST}, \ref{tab:archi CelebA} and \ref{tab:archi mobius}. In all tables, $\star$ stands for a duplicated layer in stochastic models to calculate the mean and the variance, and BN stands for batch normalization.

\begin{table}[ht]
    \centering
    \caption{Autoencoders model architecture for the MNIST data set. }
    \begin{tabular}{|c|c|c|}\toprule
        \textbf{Layers} & \textbf{Encoder} & \textbf{Decoder} \\\midrule
        Layer 1 & $\conv(128, (4,4), s=2, p=1)$ & $\lin(32, 16384)$\\
        & BN, ReLU & $\resh(1024,4,4)$\\\midrule
        Layer 2 & $\conv(256, (4,4), s=2, p=1)$ & $\convT(512, (4, 4), s=2, p=1)$\\
        & BN, ReLU & BN, ReLU\\\midrule
        Layer 3 & $\conv(512, (4,4), s=2, p=1)$ & $\convT(256, (4, 4), s=2, p=2)$\\
        & BN, ReLU & BN, ReLU\\\midrule
        Layer 4 & $\conv(1024, (4,4), s=2, p=1)$ & $\convT(256, (4, 4), s=2, p=1)$\\
        & BN, ReLU & Sigmoid\\\midrule
        Layer 5 & $\lin(1024,32)\star$ & -\\
        &  & \\\bottomrule
    \end{tabular}
    
    \label{tab:archi MNIST}
\end{table}

\begin{table}[ht]
    \centering
    \caption{Autoencoders model architecture for the CelebA data set. }
    \begin{tabular}{|c|c|c|}\toprule
        \textbf{Layers} & \textbf{Encoder} & \textbf{Decoder} \\\midrule
        Layer 1 & $\conv(128, (4,4), s=2, p=1)$ & $\lin(256, 146432)$\\
        & BN, ReLU & BN, ReLU, $\resh(1024,4,4)$\\\midrule
        Layer 2 & $\conv(256, (4,4), s=2, p=1)$ & $\convT(512, (3, 4), s=2, p=(0,1))$\\
        & BN, ReLU & BN, ReLU\\\midrule
        Layer 3 & $\conv(512, (4,4), s=2, p=1)$ & $\convT(256, (3, 4), s=2, p=(0,1))$\\
        & BN, ReLU & BN, ReLU\\\midrule
        Layer 4 & $\conv(1024, (4,4), s=2, p=1)$ & $\convT(128, (3, 4), s=2, p=1)$\\
        & BN, ReLU & BN, ReLU\\\midrule
        Layer 5 & $\lin(146432,256)\star$ & $\convT(3, (2, 4), s=2, p=0)$\\
        &  & Sigmoid \\\bottomrule
    \end{tabular}
    
    \label{tab:archi CelebA}
\end{table}

\begin{table}[ht]
    \centering
    \caption{Autoencoders model architecture for the MOBIUS data set.}
    \begin{tabular}{|c|c|c|}\toprule
        \textbf{Layers} & \textbf{Encoder} & \textbf{Decoder} \\\midrule
        Layer 1 & $\conv(32, (7,7), s=3, p=0)$ & $\lin(256, 3072)$\\
        & BN, ReLU & BN, ReLU, $\resh(1024,1,3)$\\\midrule
        Layer 2 & $\conv(64, (7,7), s=3, p=0)$ & $\convT(512, (7, 7), s=2, p=(0,0))$\\
        & BN, ReLU & BN, ReLU\\\midrule
        Layer 3 & $\conv(128, (7,7), s=3, p=0)$ & $\convT(256, (5, 7), s=3, p=(1,0))$\\
        & BN, ReLU & BN, ReLU\\\midrule
        Layer 4 & $\conv(256, (7,7), s=3, p=0)$ & $\convT(128, (5, 7), s=3, p=(1,2))$\\
        & BN, ReLU & BN, ReLU\\\midrule
        Layer 5 & $\conv(512, (7,7), s=3, p=0)$ & $\convT(64, (5, 7), s=3, p=(1,2))$\\
        & BN, ReLU & BN, ReLU\\\midrule
        Layer 6 & $\conv(1024, (5,5), s=2, p=0)$ & $\convT(32, (5, 7), s=3, p=(1,2))$\\
        & BN, ReLU & BN, ReLU\\\midrule
        Layer 7 & $\lin(3072,256)\star$ & $\convT(3, (4, 6), s=3, p=(1,0))$\\
        &  & Sigmoid \\\bottomrule
    \end{tabular}
    \label{tab:archi mobius}
\end{table}

\section{Image reconstruction}
\label{image reconstruction}
\subsection{MNIST}
\begin{figure}[H]
    \begin{subfigure} {\textwidth}
         \centering
         \includegraphics[width=.9\textwidth]{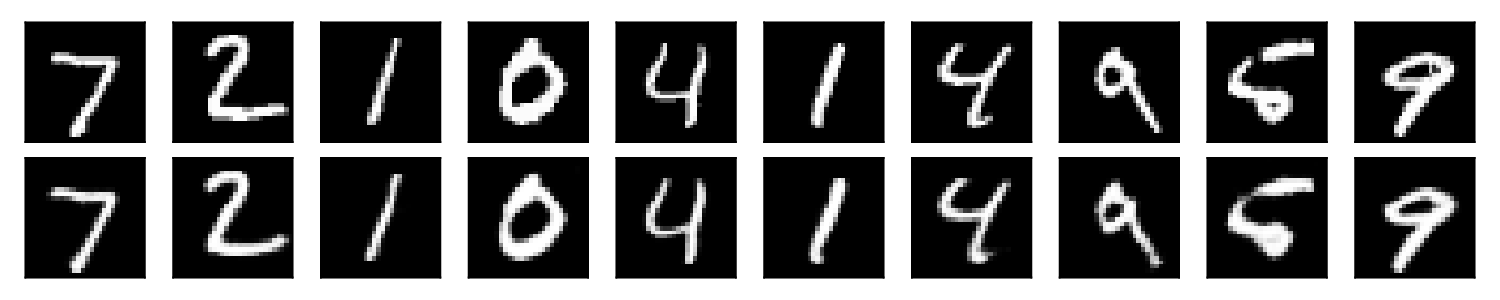}
         \caption{Vanilla Autoencoder.}
     \end{subfigure}
     \hfill
     \begin{subfigure} {\textwidth}
         \centering
         \includegraphics[width=.9\textwidth]{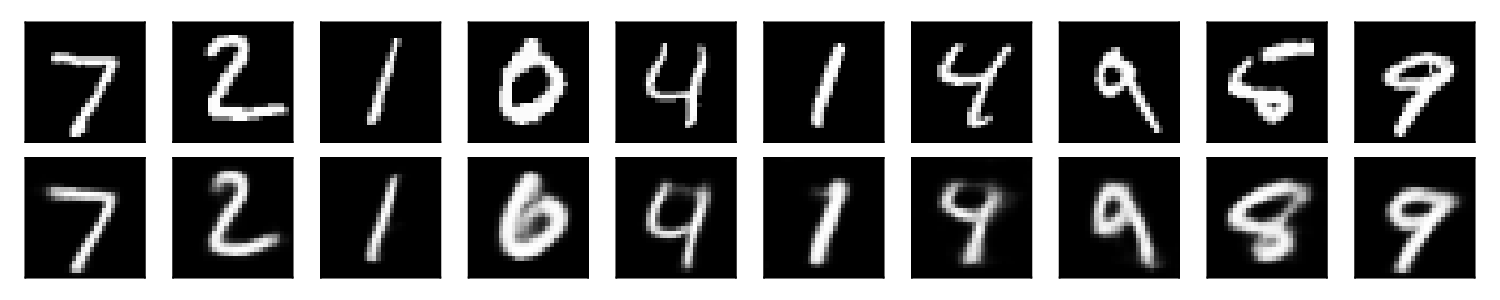}
         \caption{VAE.}
     \end{subfigure}
     \hfill
     \begin{subfigure} {\textwidth}
         \centering
         \includegraphics[width=.9\textwidth]{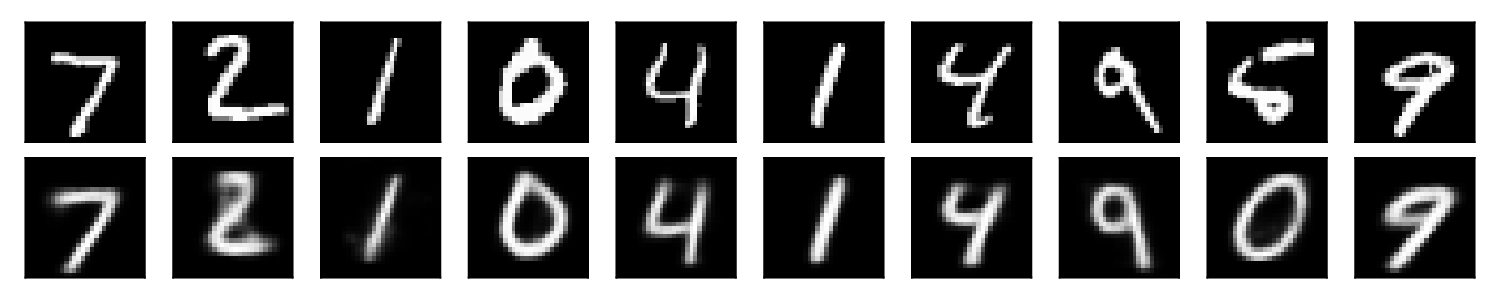}
         \caption{$\beta$-VAE.}
     \end{subfigure}
     \hfill
     \begin{subfigure} {\textwidth}
         \centering
         \includegraphics[width=.9\textwidth]{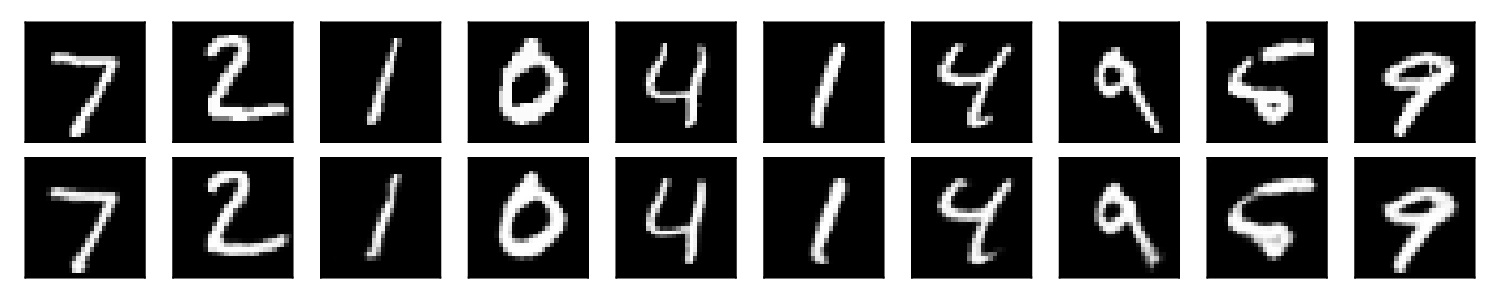}
         \caption{WAE.}
     \end{subfigure}
     \hfill
     \begin{subfigure} {\textwidth}
         \centering
         \includegraphics[width=.9\textwidth]{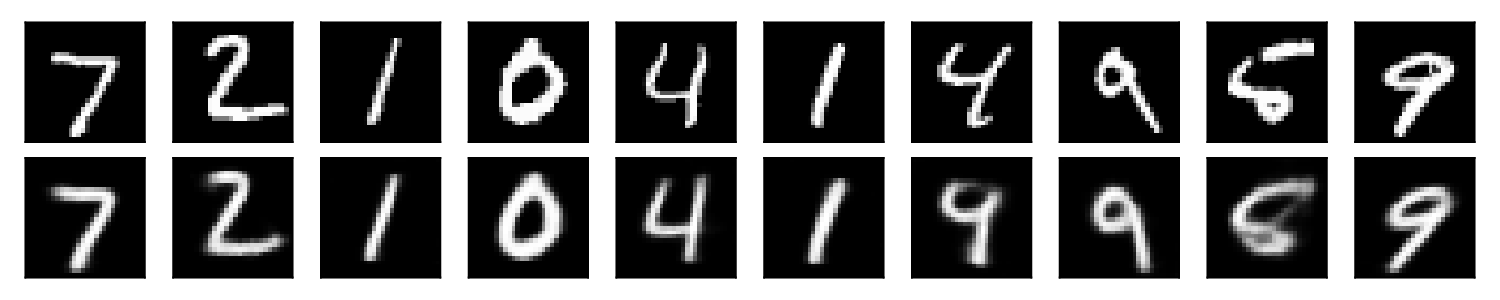}
         \caption{InfoVAE.}
     \end{subfigure}
     \caption{Reconstruction of MNIST images via different autoencoder based models.}
\end{figure}

\newpage
\subsection{CelebA}
\begin{figure}[H]
    \begin{subfigure} {\textwidth}
         \centering
         \includegraphics[width=\textwidth]{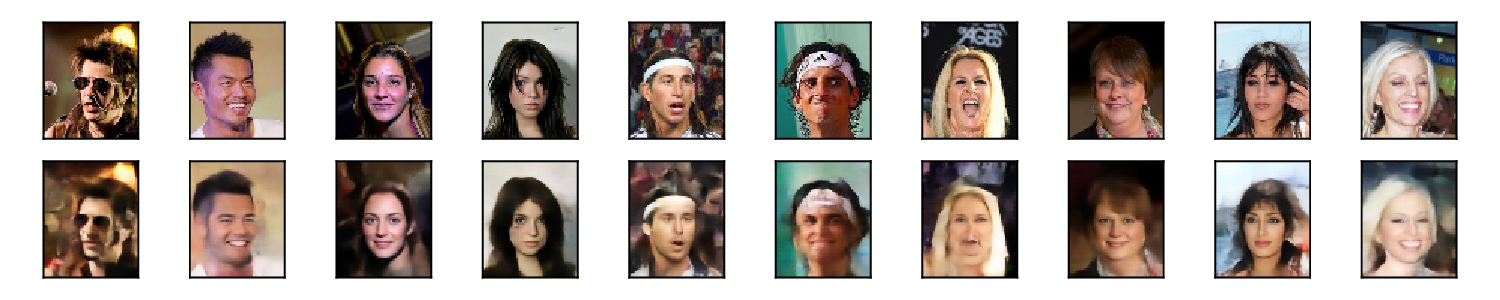}
         \caption{Vanilla Autoencoder.}
     \end{subfigure}
     \hfill
     \begin{subfigure} {\textwidth}
         \centering
         \includegraphics[width=\textwidth]{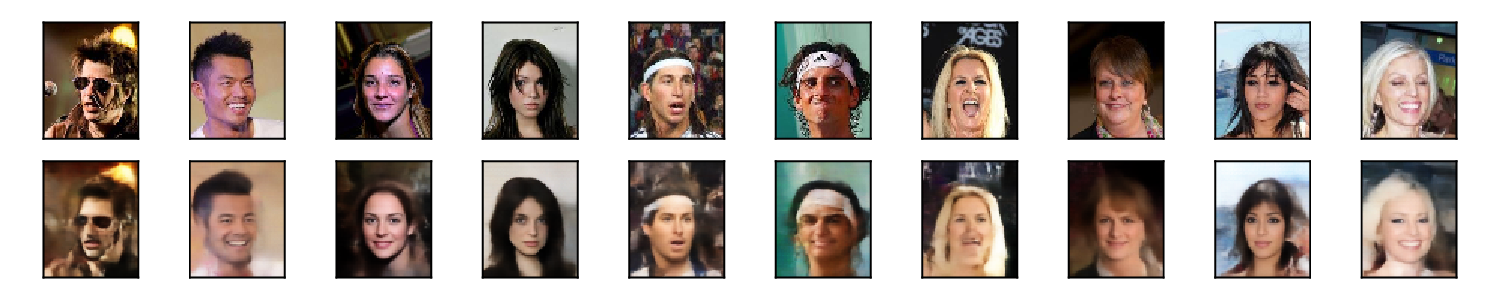}
         \caption{VAE.}
     \end{subfigure}
     \hfill
     \begin{subfigure} {\textwidth}
         \centering
         \includegraphics[width=\textwidth]{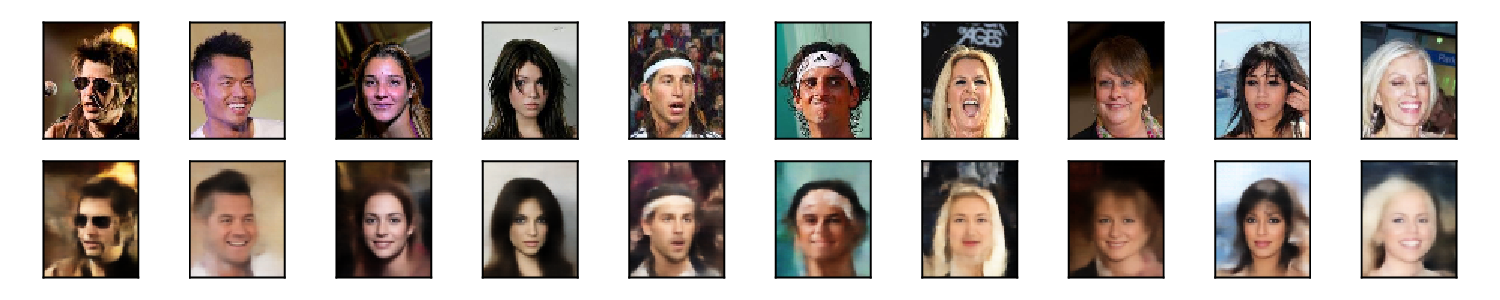}
         \caption{$\beta$-VAE.}
     \end{subfigure}
     \hfill
     \begin{subfigure} {\textwidth}
         \centering
         \includegraphics[width=\textwidth]{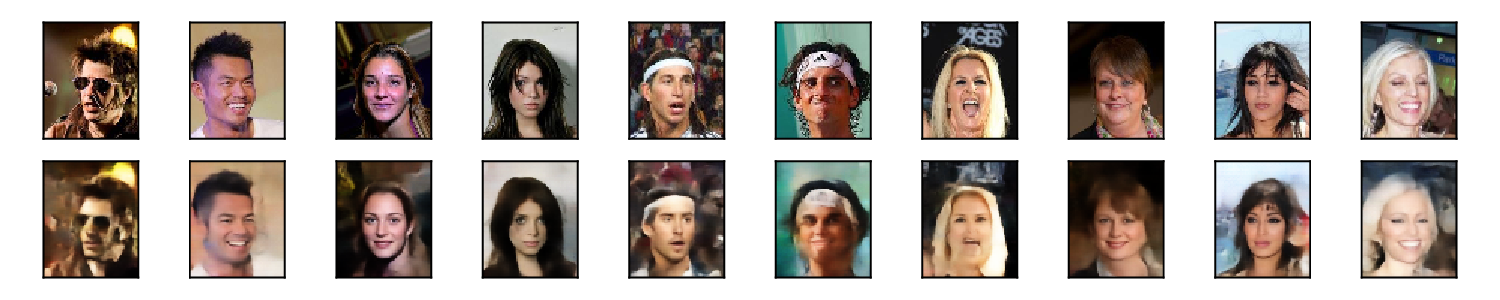}
         \caption{WAE.}
     \end{subfigure}
     \hfill
     \begin{subfigure} {\textwidth}
         \centering
         \includegraphics[width=\textwidth]{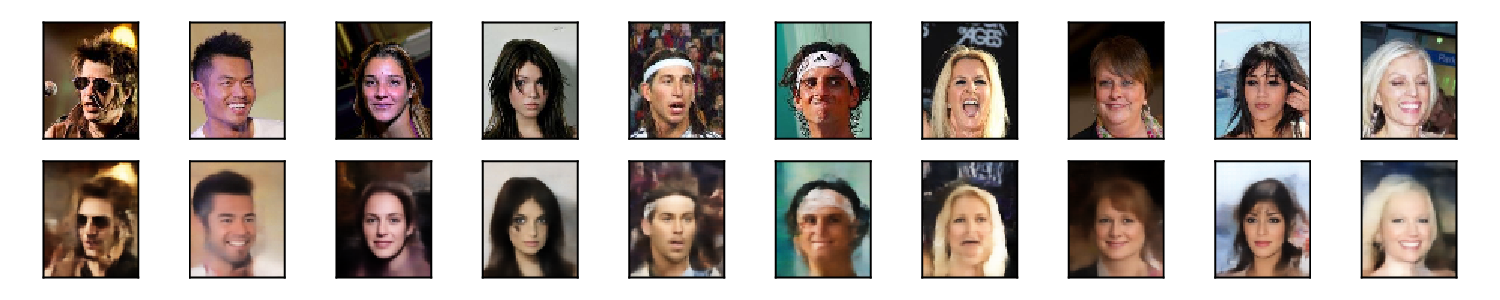}
         \caption{InfoVAE.}
     \end{subfigure}
     \caption{Reconstruction of CelebA images via different autoencoder based models.}
\end{figure}

\newpage
\subsection{MOBIUS}
\begin{figure}[H]
    \begin{subfigure} {\textwidth}
         \centering
         \includegraphics[width=.9\textwidth]{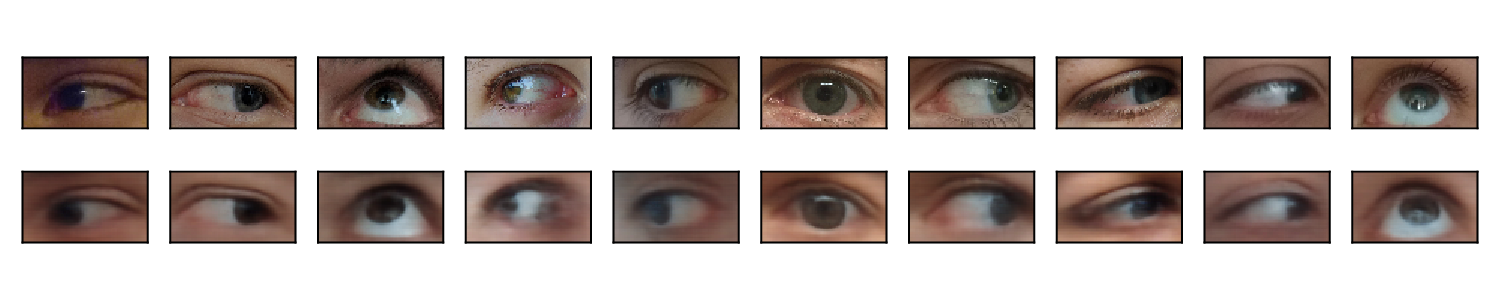}
         \caption{Vanilla Autoencoder.}
     \end{subfigure}
     \hfill
     \begin{subfigure} {\textwidth}
         \centering
         \includegraphics[width=.9\textwidth]{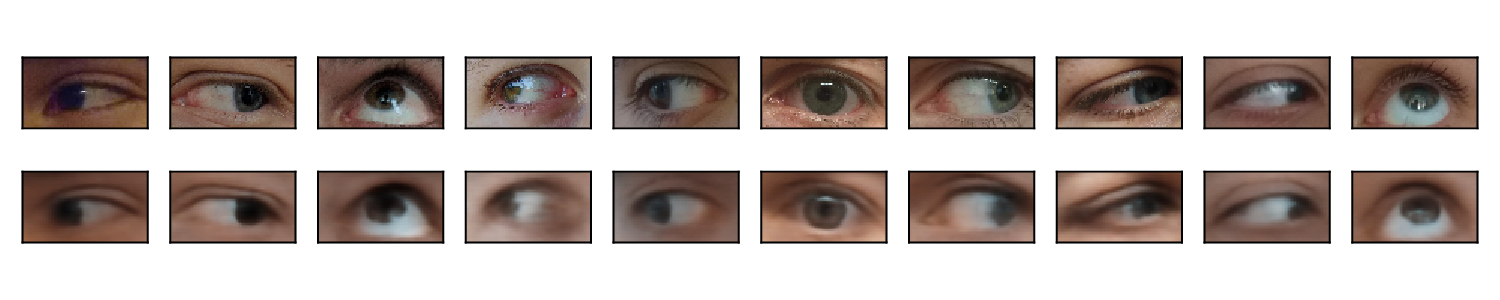}
         \caption{VAE.}
     \end{subfigure}
     \hfill
     \begin{subfigure} {\textwidth}
         \centering
         \includegraphics[width=.9\textwidth]{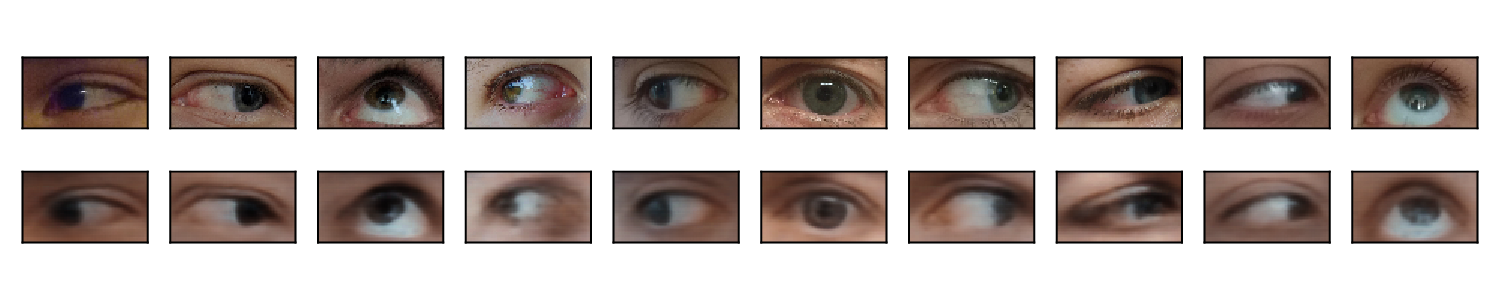}
         \caption{$\beta$-VAE.}
     \end{subfigure}
     \hfill
     \begin{subfigure} {\textwidth}
         \centering
         \includegraphics[width=.9\textwidth]{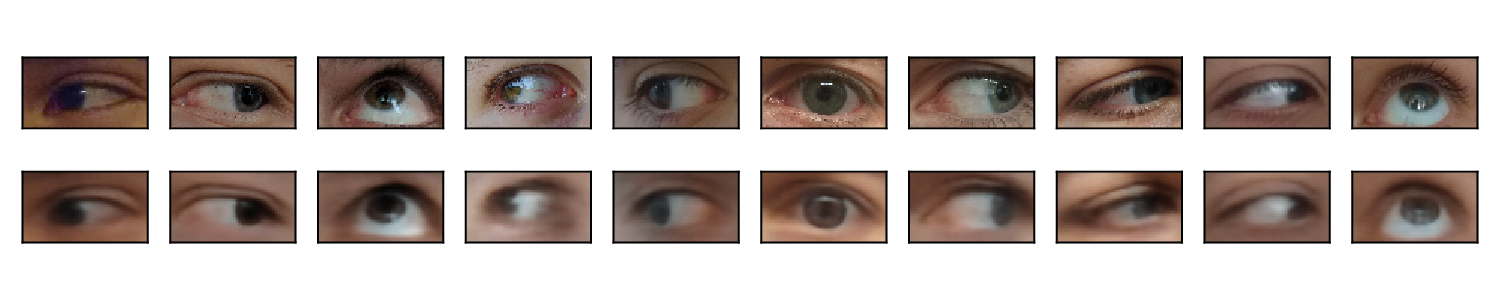}
         \caption{WAE.}
     \end{subfigure}
     \hfill
     \begin{subfigure} {\textwidth}
         \centering
         \includegraphics[width=.9\textwidth]{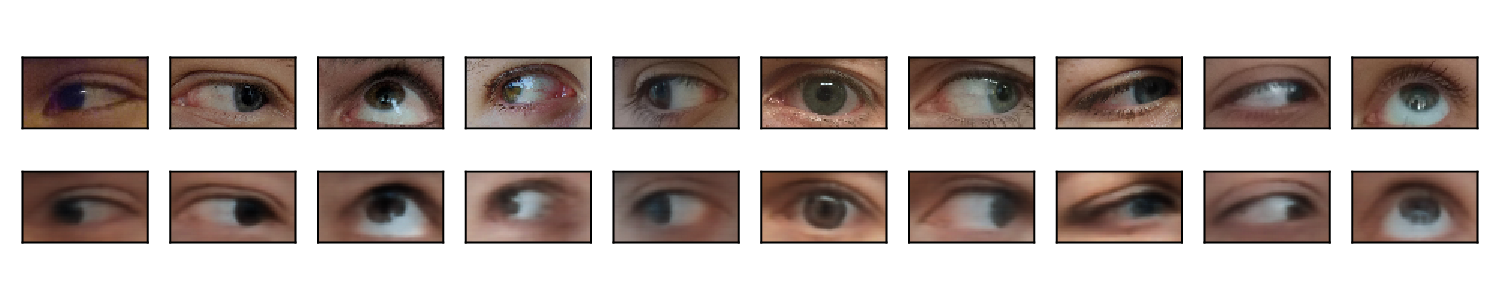}
         \caption{InfoVAE.}
     \end{subfigure}
     \caption{Reconstruction of MOBIUS images via different autoencoder based models.}
\end{figure}

\newpage
\section{Synthetic images}
\label{synthetic images}
\subsection{MNIST}
\begin{figure}[H]
    \begin{subfigure} {\textwidth}
         \centering
         \includegraphics[width=.9\textwidth]{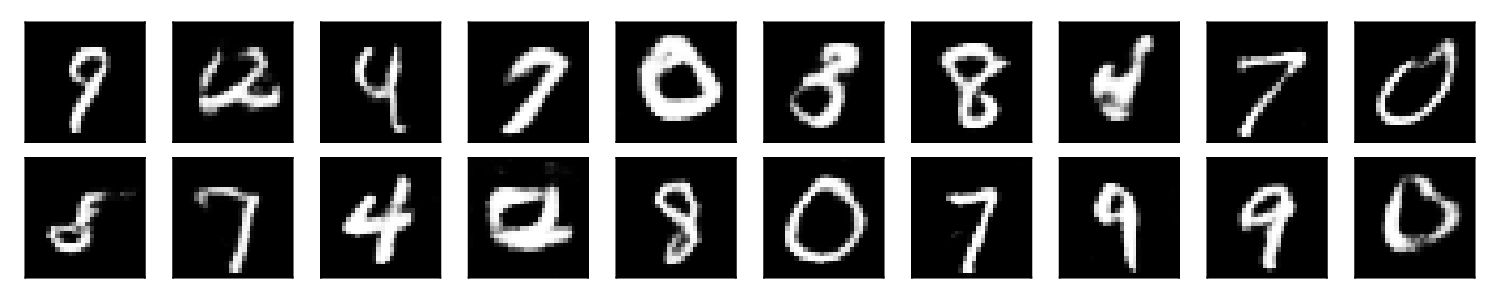}
         \caption{GMM Vanilla Autoencoder.}
     \end{subfigure}
     \hfill
     \begin{subfigure} {\textwidth}
         \centering
         \includegraphics[width=.9\textwidth]{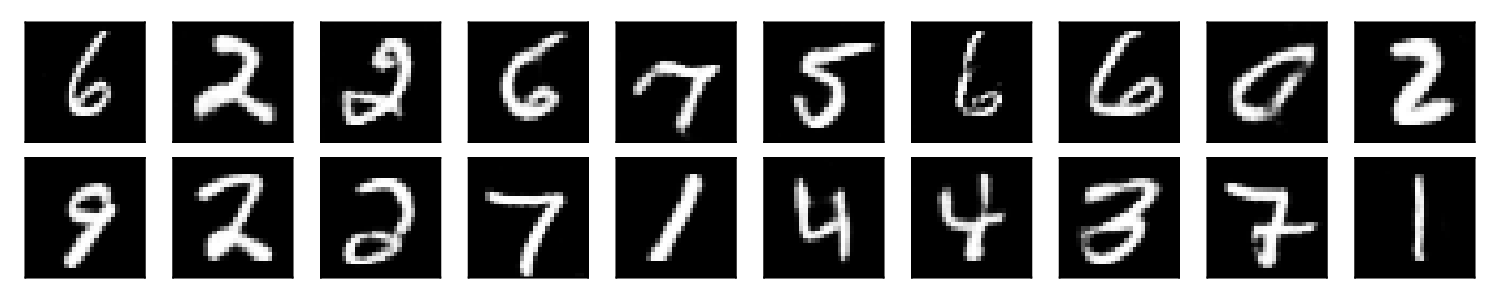}
         \caption{PMFS Vanilla Autoencoder.}
     \end{subfigure}
     \hfill
     \begin{subfigure} {\textwidth}
         \centering
         \includegraphics[width=.9\textwidth]{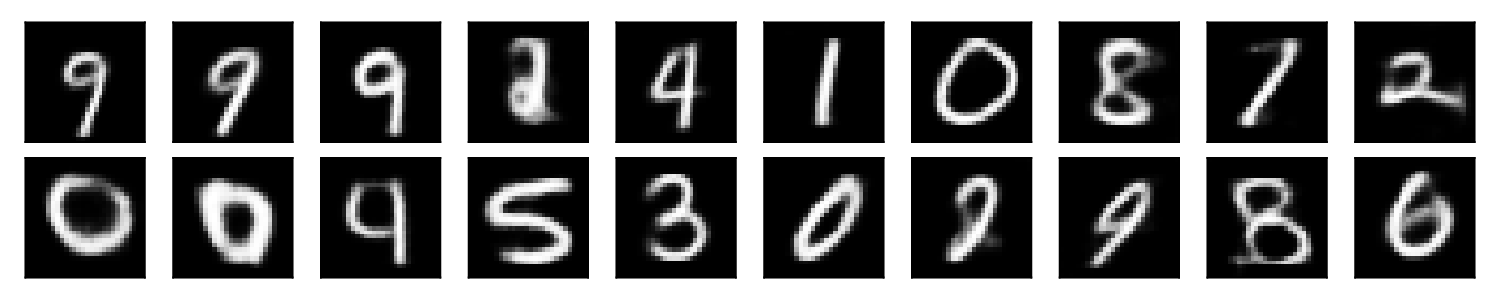}
         \caption{GMM VAE.}
     \end{subfigure}
     \hfill
     \begin{subfigure} {\textwidth}
         \centering
         \includegraphics[width=.9\textwidth]{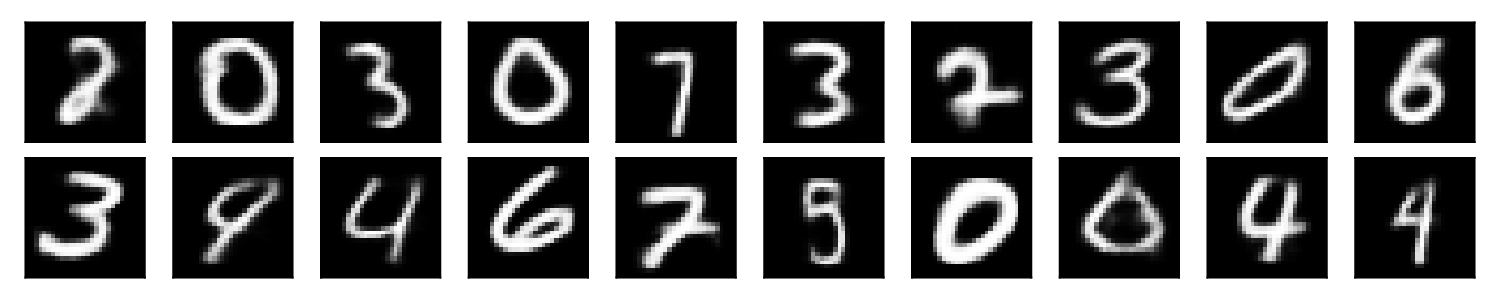}
         \caption{PMFS VAE.}
     \end{subfigure}
     \hfill
     \begin{subfigure} {\textwidth}
         \centering
         \includegraphics[width=.9\textwidth]{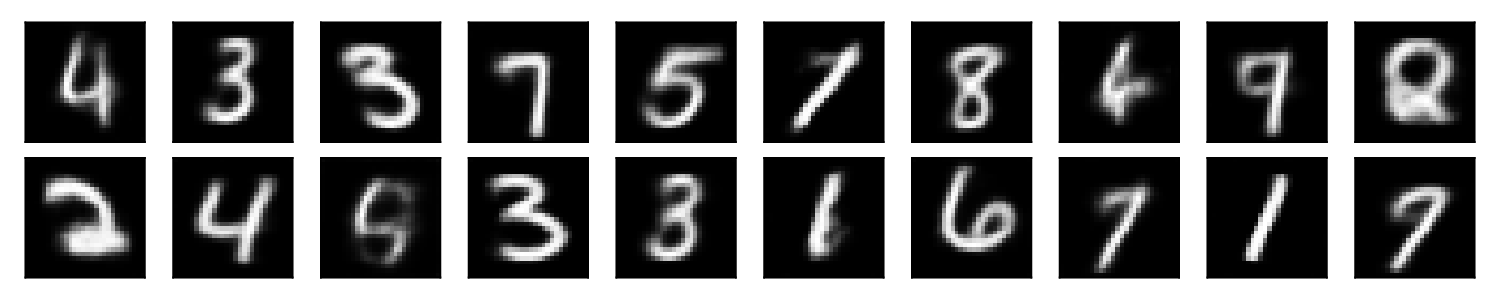}
         \caption{GMM $\beta$-VAE.}
     \end{subfigure}
\end{figure}
\begin{figure}[H]\ContinuedFloat
     \begin{subfigure} {\textwidth}
         \centering
         \includegraphics[width=.9\textwidth]{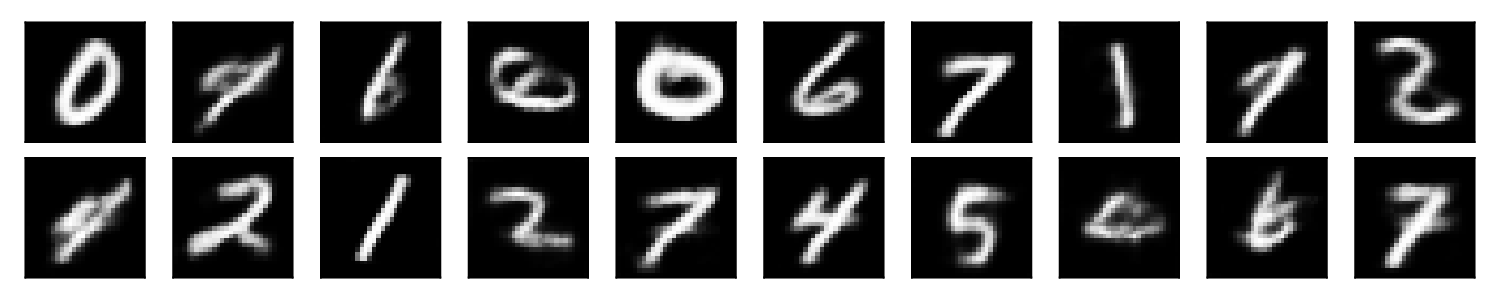}
         \caption{PMFS $\beta$-VAE.}
     \end{subfigure}
     \hfill
     \begin{subfigure} {\textwidth}
         \centering
         \includegraphics[width=.9\textwidth]{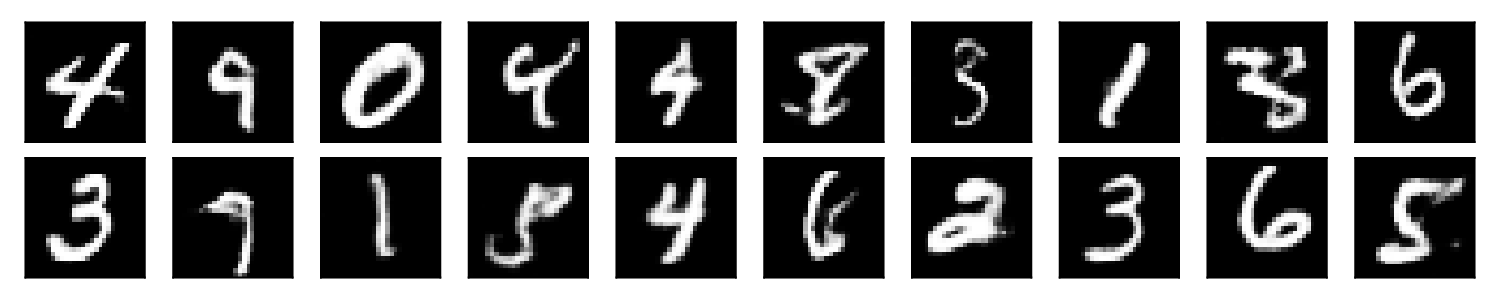}
         \caption{GMM WAE.}
     \end{subfigure}
     \hfill
     \begin{subfigure} {\textwidth}
         \centering
         \includegraphics[width=.9\textwidth]{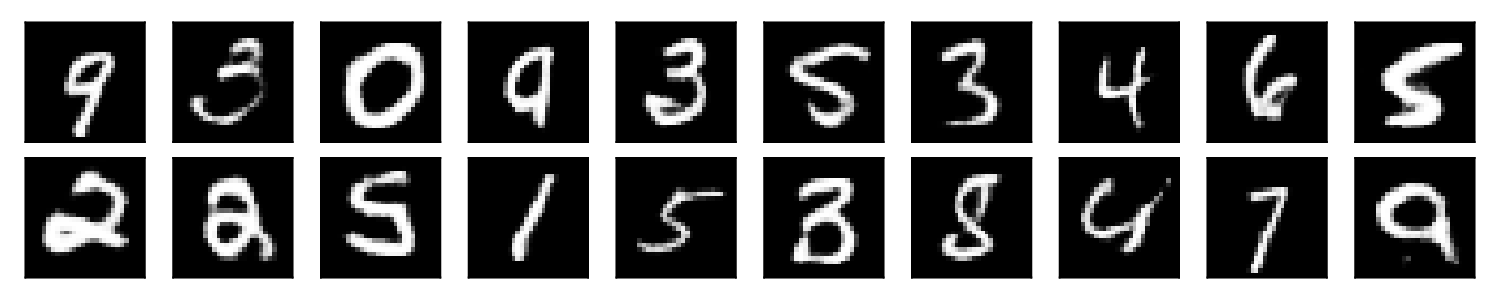}
         \caption{PMFS WAE.}
     \end{subfigure}
     \hfill
     \begin{subfigure} {\textwidth}
         \centering
         \includegraphics[width=.9\textwidth]{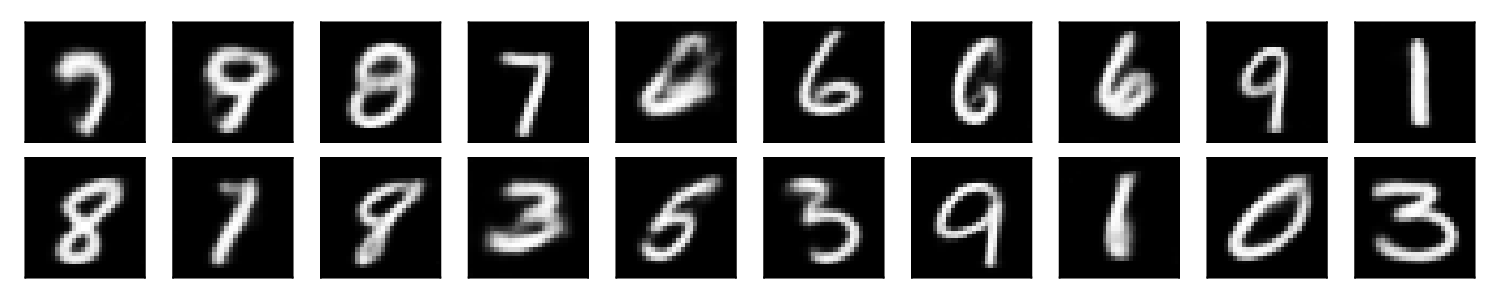}
         \caption{GMM InfoVAE.}
     \end{subfigure}
     \hfill
     \begin{subfigure} {\textwidth}
         \centering
         \includegraphics[width=.9\textwidth]{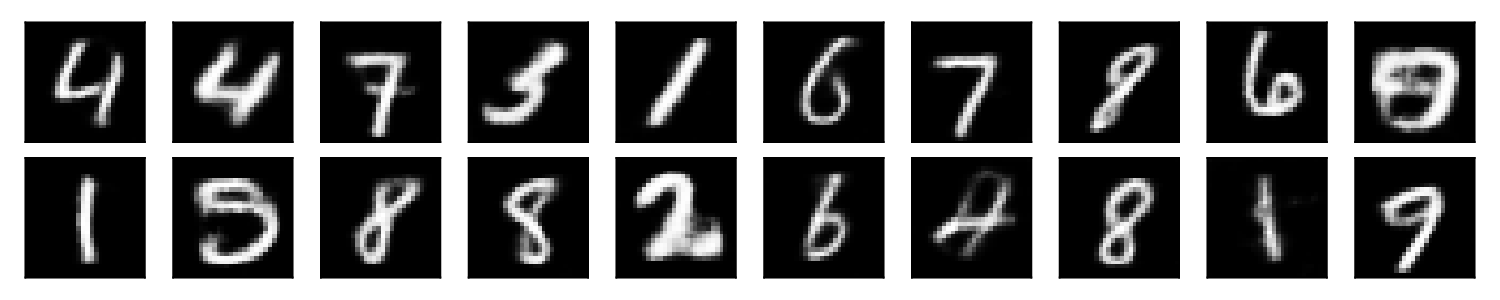}
         \caption{PMFS InfoVAE.}
     \end{subfigure}
     \caption{Synthetic images generated using GMM and PMFS sampling on the MNIST data set via different autoencoder based models.}
\end{figure}

\newpage
\subsection{CelebA}
\begin{figure}[H]
    \begin{subfigure} {\textwidth}
         \centering
         \includegraphics[width=\textwidth]{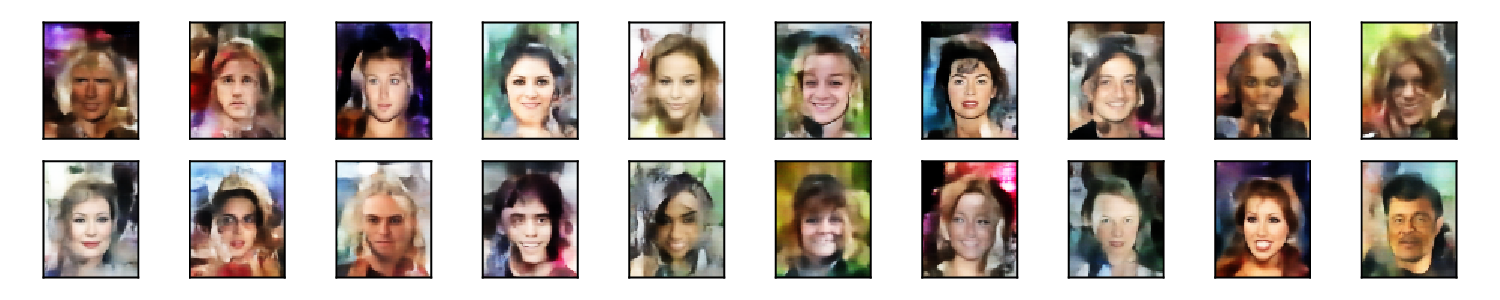}
         \caption{GMM Vanilla Autoencoder.}
     \end{subfigure}
     \hfill
     \begin{subfigure} {\textwidth}
         \centering
         \includegraphics[width=\textwidth]{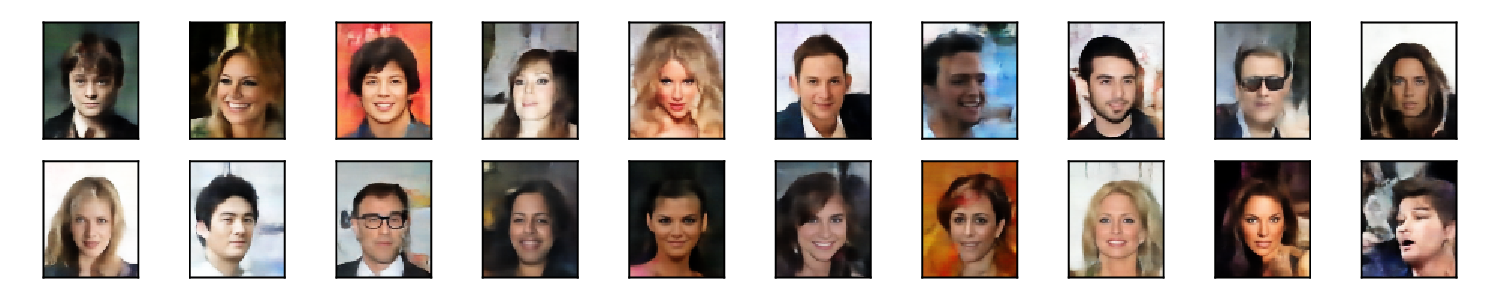}
         \caption{PMFS Vanilla Autoencoder.}
     \end{subfigure}
     \hfill
     \begin{subfigure} {\textwidth}
         \centering
         \includegraphics[width=\textwidth]{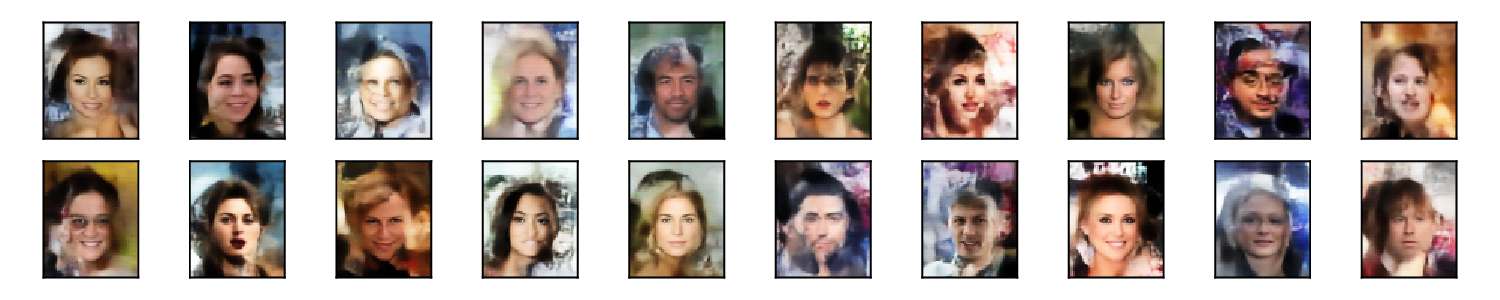}
         \caption{GMM VAE.}
     \end{subfigure}
     \hfill
     \begin{subfigure} {\textwidth}
         \centering
         \includegraphics[width=\textwidth]{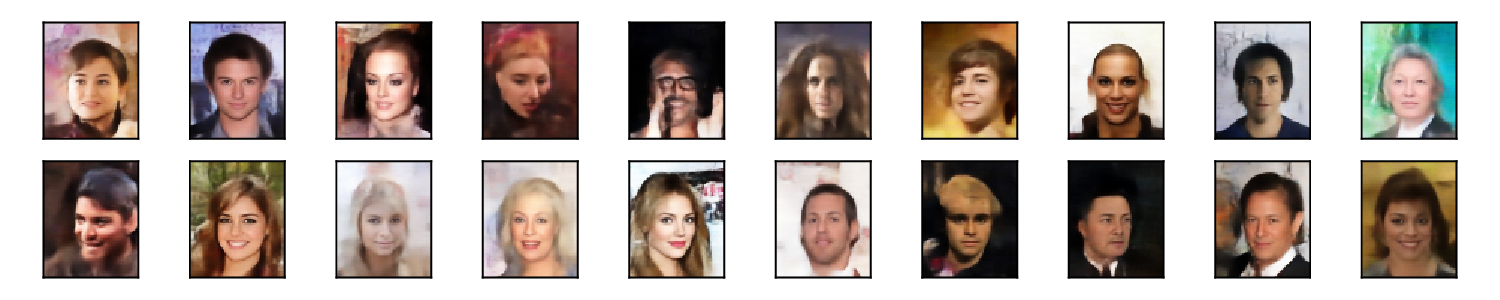}
         \caption{PMFS VAE.}
     \end{subfigure}
     \hfill
     \begin{subfigure} {\textwidth}
         \centering
         \includegraphics[width=\textwidth]{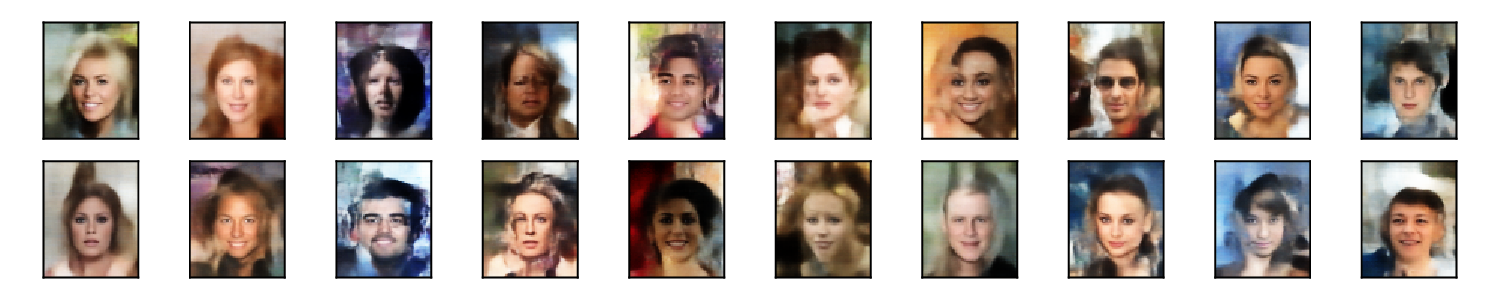}
         \caption{GMM $\beta$-VAE.}
     \end{subfigure}
\end{figure}
\begin{figure}[H]\ContinuedFloat
     \begin{subfigure} {\textwidth}
         \centering
         \includegraphics[width=\textwidth]{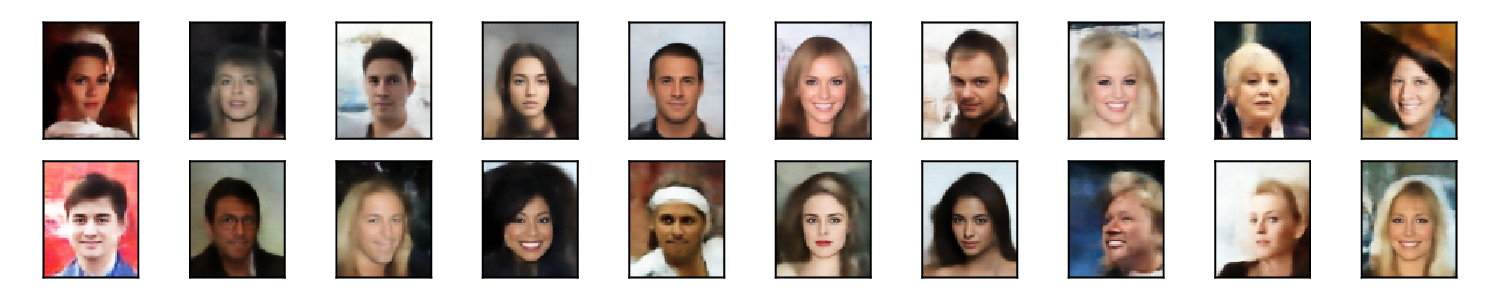}
         \caption{PMFS $\beta$-VAE.}
     \end{subfigure}
     \hfill
     \begin{subfigure} {\textwidth}
         \centering
         \includegraphics[width=\textwidth]{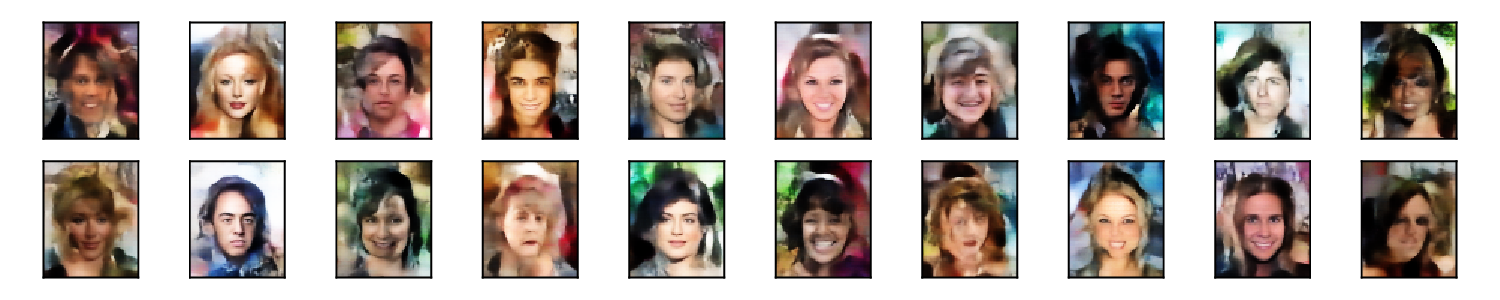}
         \caption{GMM WAE.}
     \end{subfigure}
     \hfill
     \begin{subfigure} {\textwidth}
         \centering
         \includegraphics[width=\textwidth]{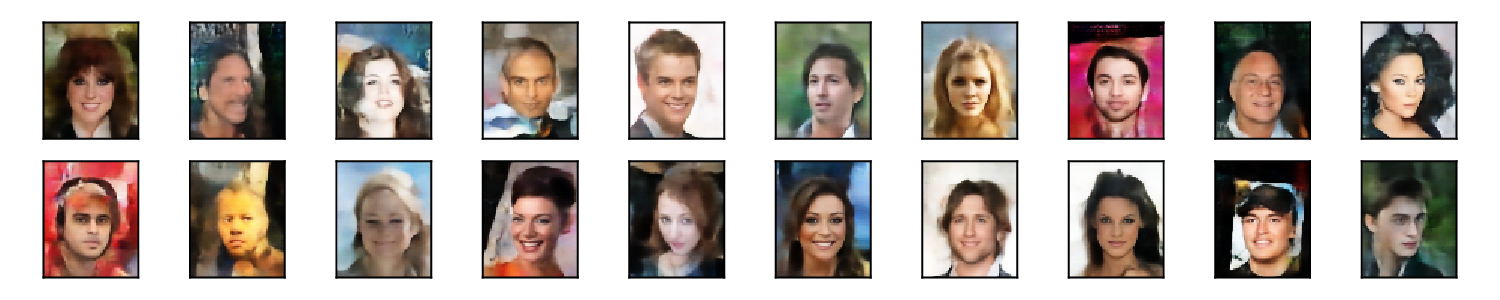}
         \caption{PMFS WAE.}
     \end{subfigure}
     \hfill
     \begin{subfigure} {\textwidth}
         \centering
         \includegraphics[width=\textwidth]{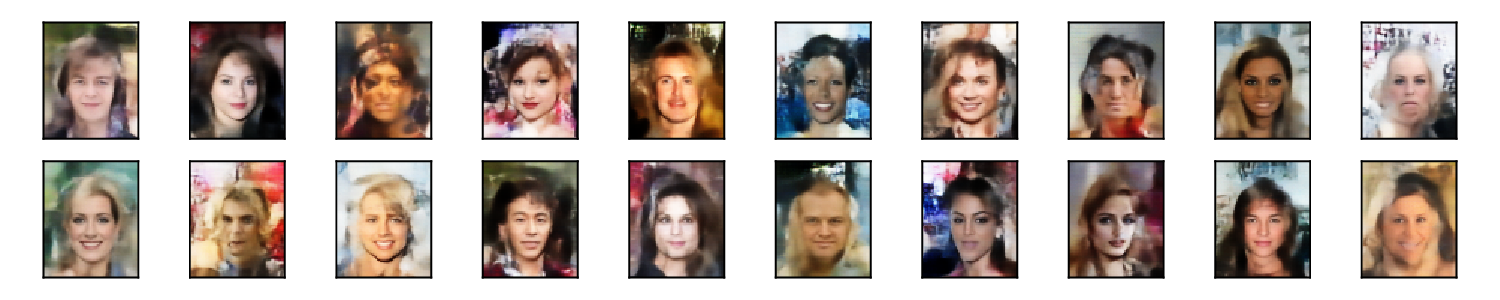}
         \caption{GMM InfoVAE.}
     \end{subfigure}
     \hfill
     \begin{subfigure} {\textwidth}
         \centering
         \includegraphics[width=\textwidth]{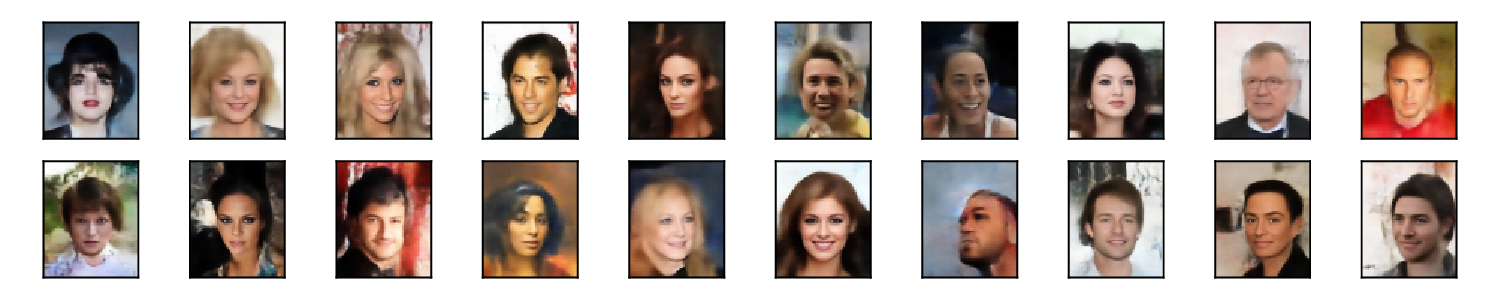}
         \caption{PMFS InfoVAE.}
     \end{subfigure}
     \caption{Synthetic images generated using GMM and PMFS sampling on the CelebA data set via different autoencoder based models.}
\end{figure}

\newpage
\subsection{MOBIUS}
\begin{figure}[H]
    \begin{subfigure} {\textwidth}
         \centering
         \includegraphics[width=.9\textwidth]{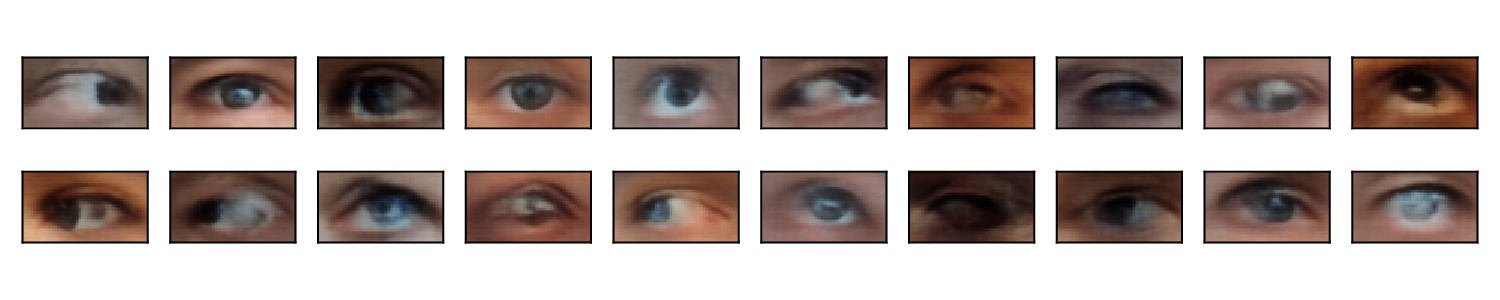}
         \caption{GMM Vanilla Autoencoder.}
     \end{subfigure}
     \hfill
     \begin{subfigure} {\textwidth}
         \centering
         \includegraphics[width=.9\textwidth]{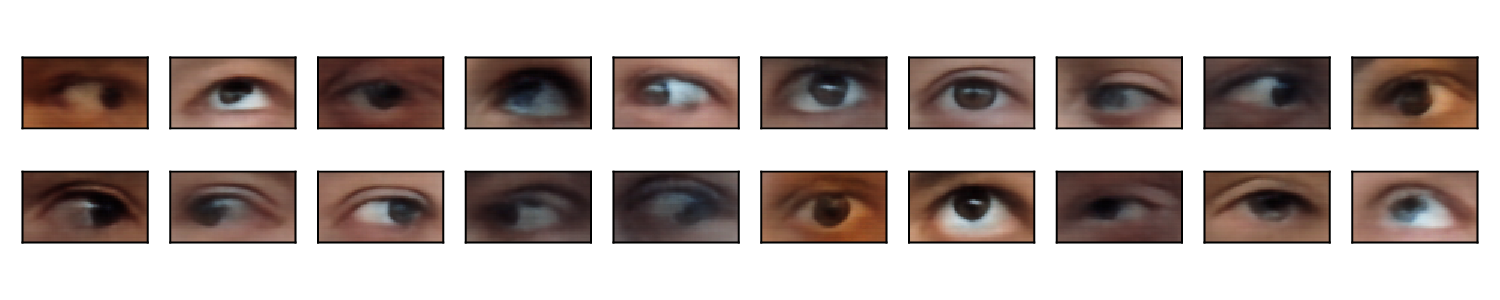}
         \caption{PMFS Vanilla Autoencoder.}
     \end{subfigure}
     \hfill
     \begin{subfigure} {\textwidth}
         \centering
         \includegraphics[width=.9\textwidth]{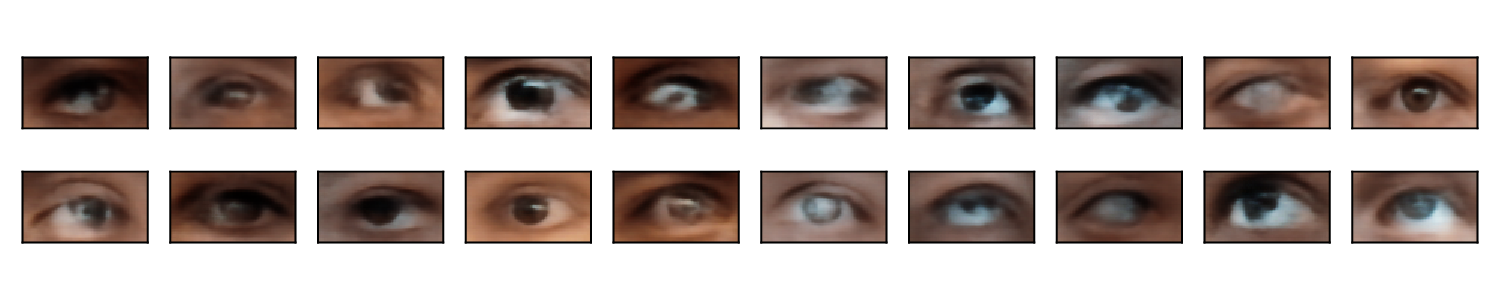}
         \caption{GMM VAE.}
     \end{subfigure}
     \hfill
     \begin{subfigure} {\textwidth}
         \centering
         \includegraphics[width=.9\textwidth]{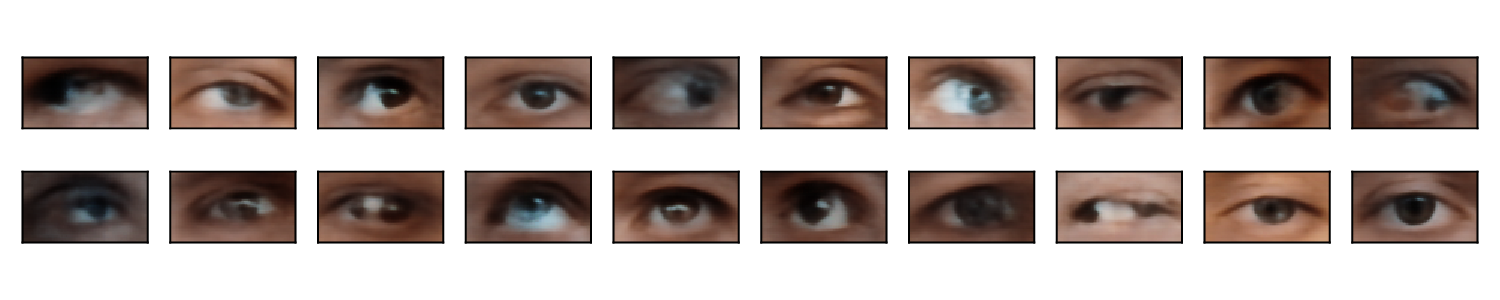}
         \caption{PMFS VAE.}
     \end{subfigure}
     \hfill
     \begin{subfigure} {\textwidth}
         \centering
         \includegraphics[width=.9\textwidth]{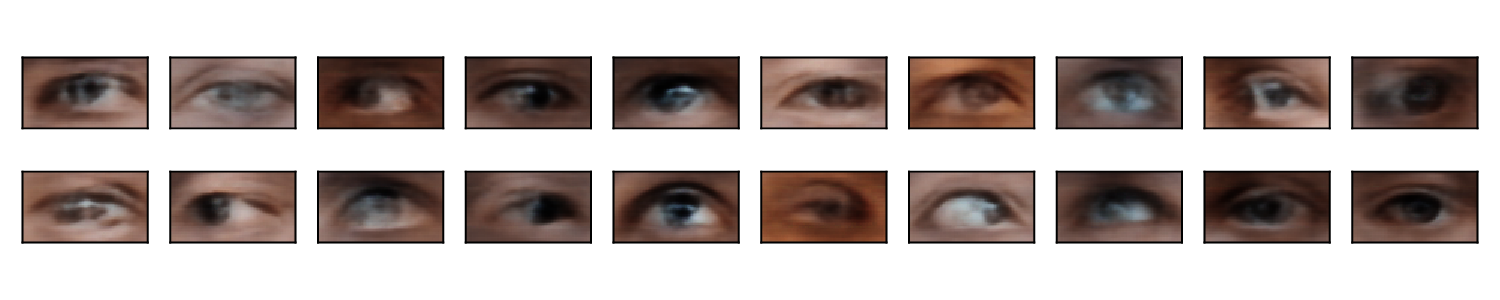}
         \caption{GMM $\beta$-VAE.}
     \end{subfigure}
\end{figure}
\begin{figure}[H]\ContinuedFloat
     \begin{subfigure} {\textwidth}
         \centering
         \includegraphics[width=.9\textwidth]{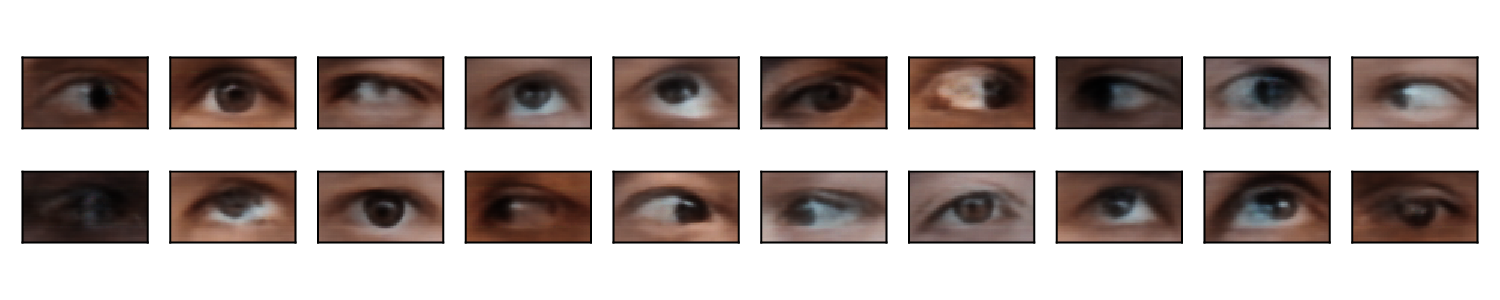}
         \caption{PMFS $\beta$-VAE.}
     \end{subfigure}
     \hfill
     \begin{subfigure} {\textwidth}
         \centering
         \includegraphics[width=.9\textwidth]{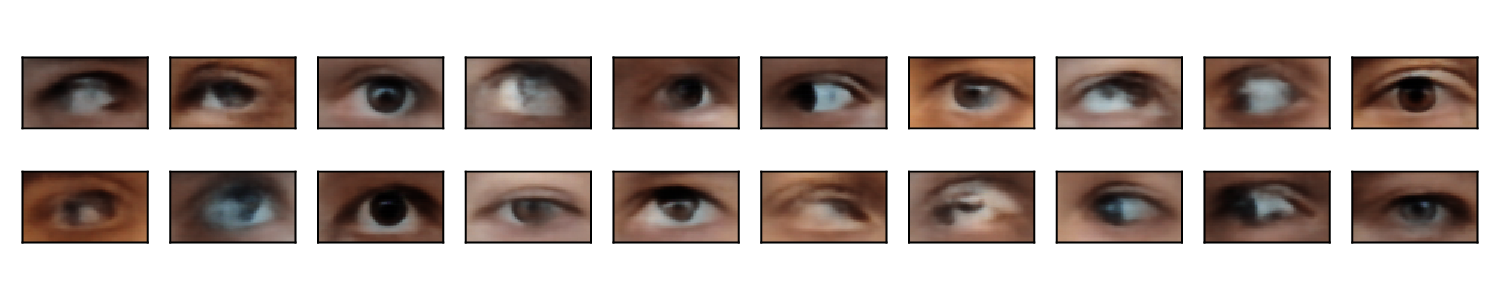}
         \caption{GMM WAE.}
     \end{subfigure}
     \hfill
     \begin{subfigure} {\textwidth}
         \centering
         \includegraphics[width=.9\textwidth]{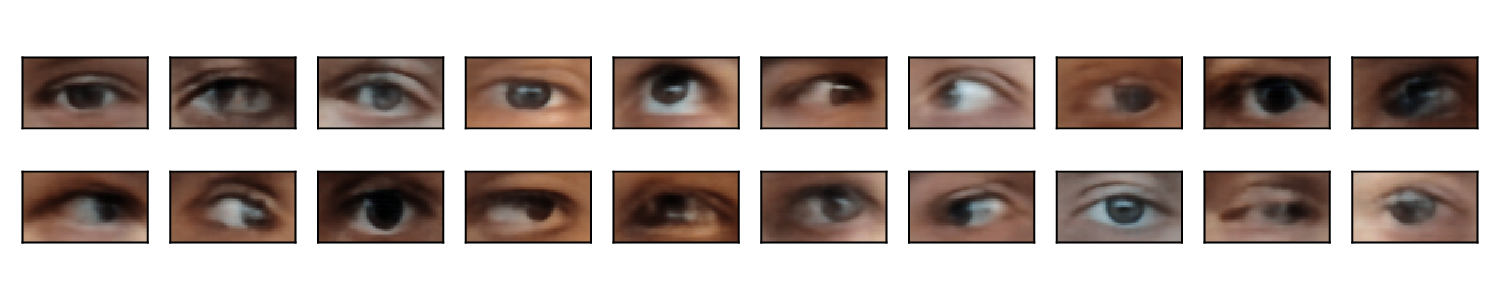}
         \caption{PMFS WAE.}
     \end{subfigure}
     \hfill
     \begin{subfigure} {\textwidth}
         \centering
         \includegraphics[width=.9\textwidth]{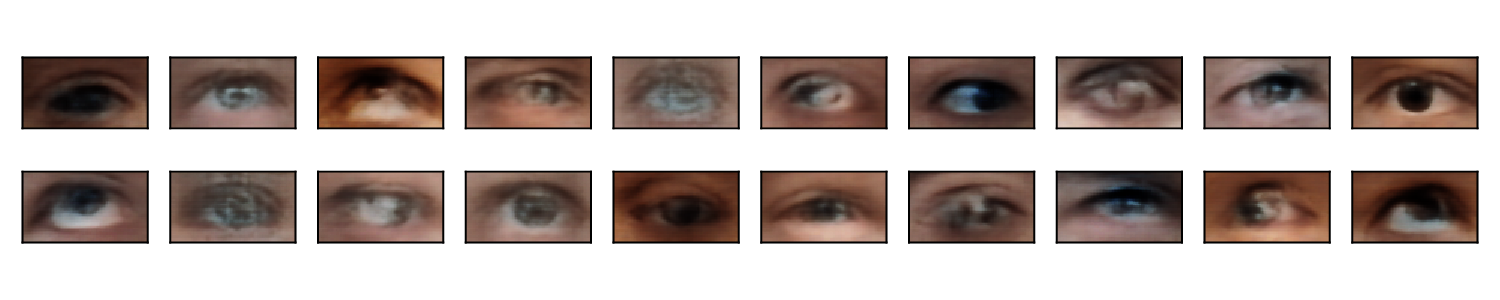}
         \caption{GMM InfoVAE.}
     \end{subfigure}
     \hfill
     \begin{subfigure} {\textwidth}
         \centering
         \includegraphics[width=.9\textwidth]{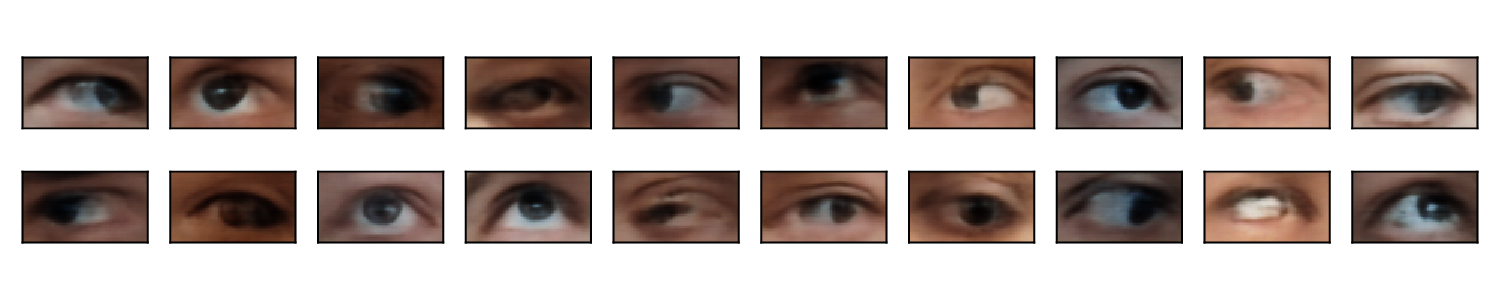}
         \caption{PMFS InfoVAE.}
     \end{subfigure}
     \caption{Synthetic images generated using GMM and PMFS sampling on the MOBIUS data set via different autoencoder based models.}
\end{figure}


\section{Distribution plots}
\label{dists}
\subsection{MNIST}

\begin{figure}[H]
    \centering
    \begin{subfigure} {0.49\textwidth}
         \centering
         \includegraphics[width=.9\textwidth]{figures/MNIST/latent_dist_ae.eps}
         \caption{Latent space distribution of Vanilla Autoencoder.}
     \end{subfigure}
     \hfill
     \begin{subfigure} {0.49\textwidth}
         \centering
         \includegraphics[width=.9\textwidth]{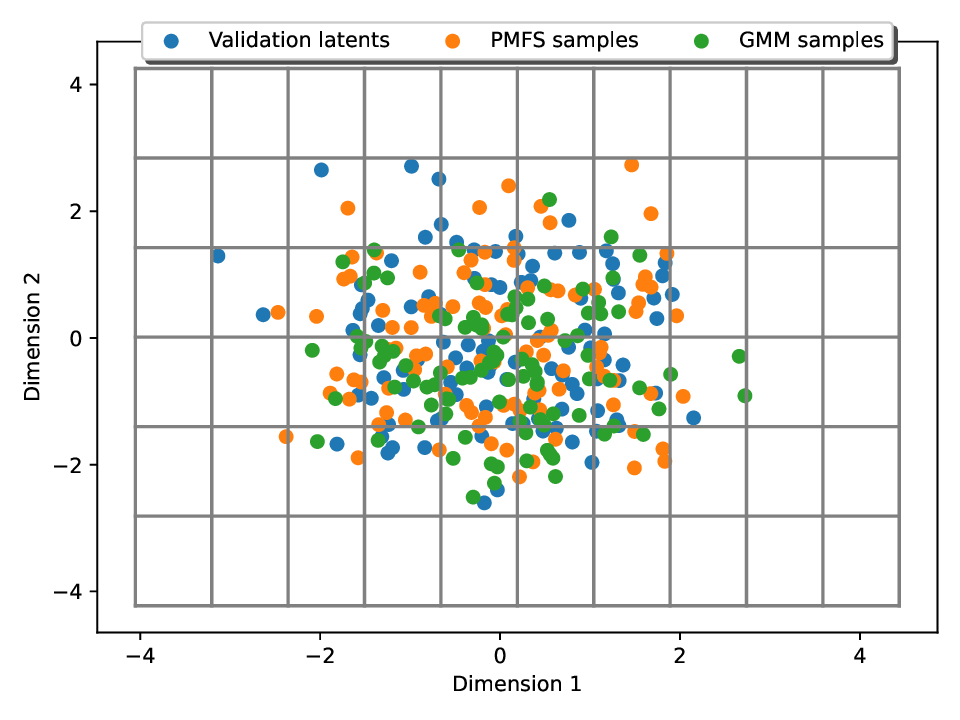}
         \caption{Latent space distribution of VAE.}
     \end{subfigure}
     \hfill
     \begin{subfigure} {0.49\textwidth}
         \centering
         \includegraphics[width=.9\textwidth]{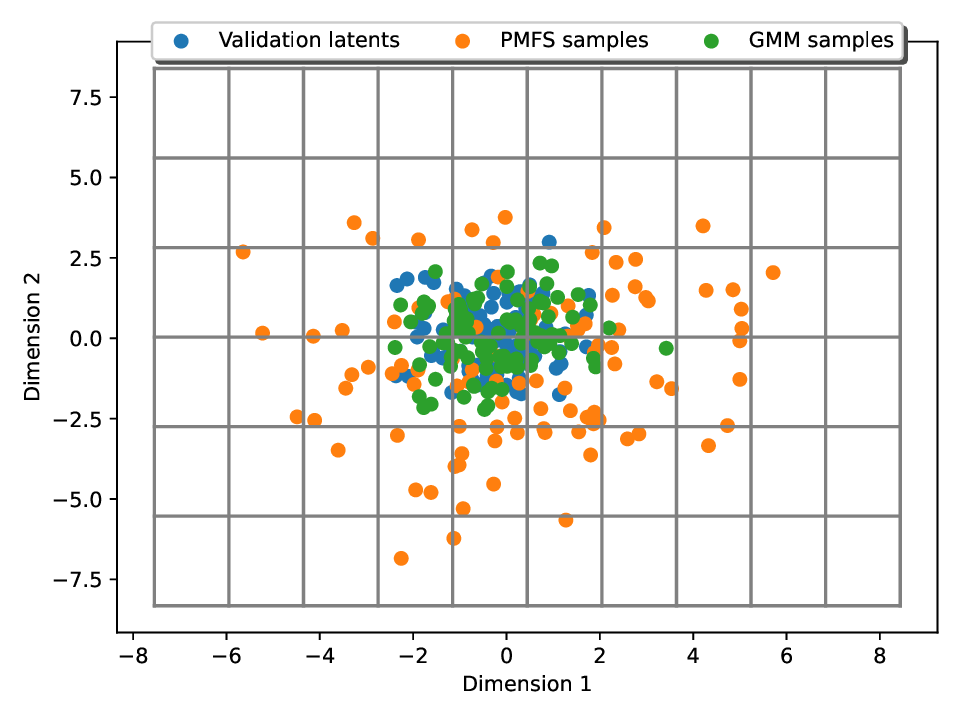}
         \caption{Latent space distribution of $\beta$-VAE.}
     \end{subfigure}
     \hfill
     \begin{subfigure} {0.49\textwidth}
         \centering
         \includegraphics[width=.9\textwidth]{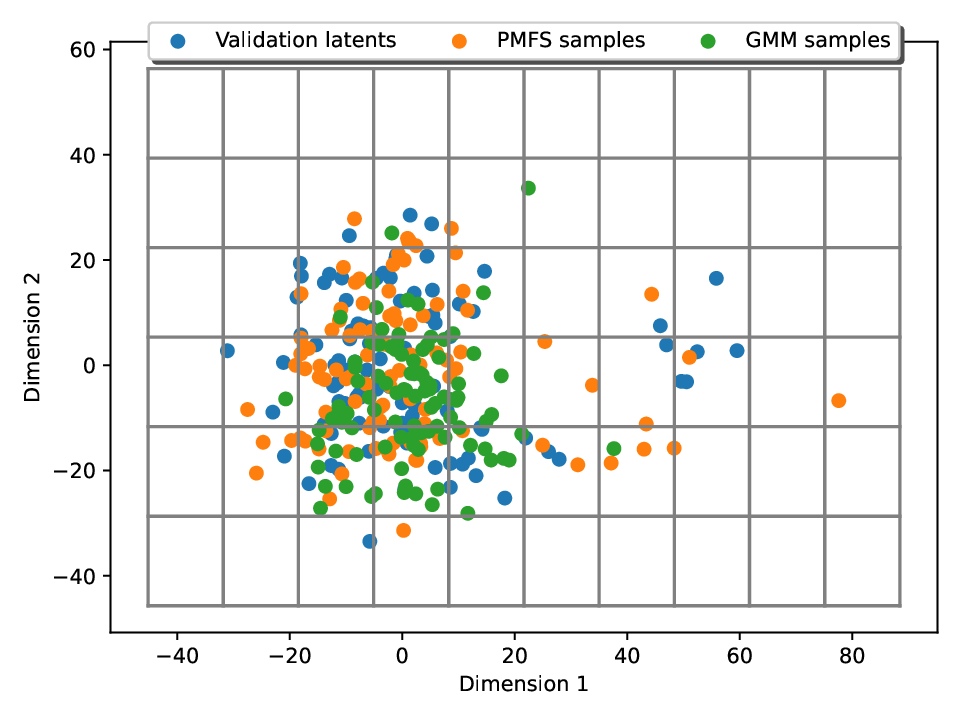}
         \caption{Latent space distribution of WAE.}
     \end{subfigure}
     \hfill
     \begin{subfigure} {0.49\textwidth}
         \centering
         \includegraphics[width=.9\textwidth]{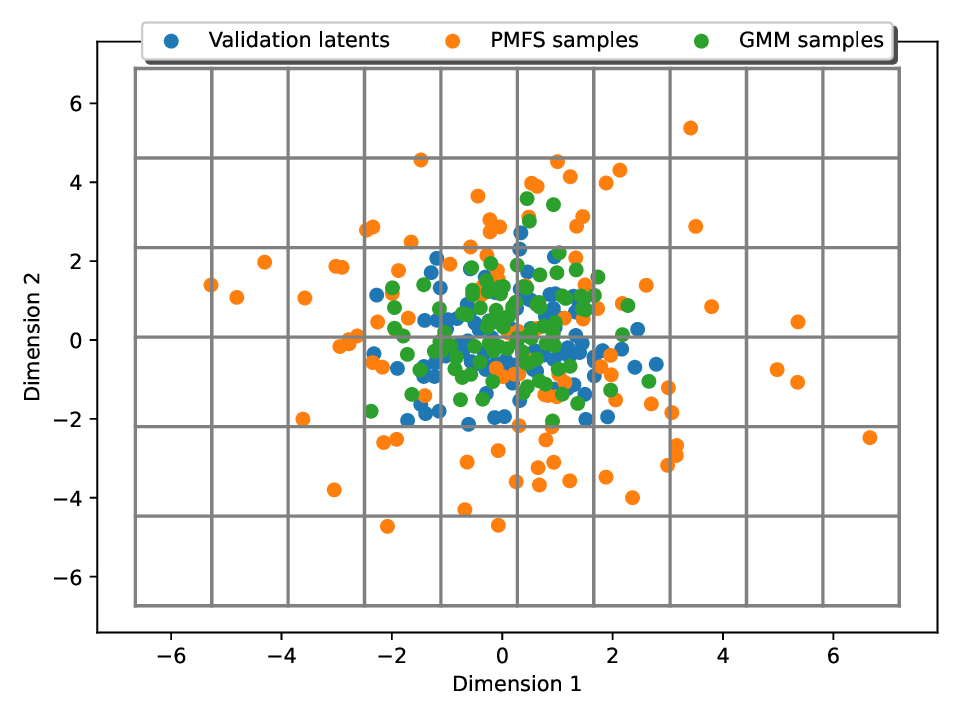}
         \caption{Latent space distribution of InfoVAE.}
     \end{subfigure}
     \hfill
     \caption{Distribution of the validation set’s latent vectors of different Autoencoder models and samples generated via GMM and PMFS sampling on the MNIST data set. These figures represents a dimensionality reduction using PCA into the two dimensional $\mathbb{R}^2$ space of the first hundred images from the validation set, as well as a hundred samples from GMM and PMFS sampling methods.}
    \label{dist_mnist}
\end{figure}

\newpage
\subsection{CelebA}
\begin{figure}[H]
    \centering
    \begin{subfigure} {0.49\textwidth}
         \centering
         \includegraphics[width=.9\textwidth]{figures/celebA/latent_dist_ae.eps}
         \caption{Latent space distribution of Vanilla Autoencoder.}
     \end{subfigure}
     \hfill
     \begin{subfigure} {0.49\textwidth}
         \centering
         \includegraphics[width=.9\textwidth]{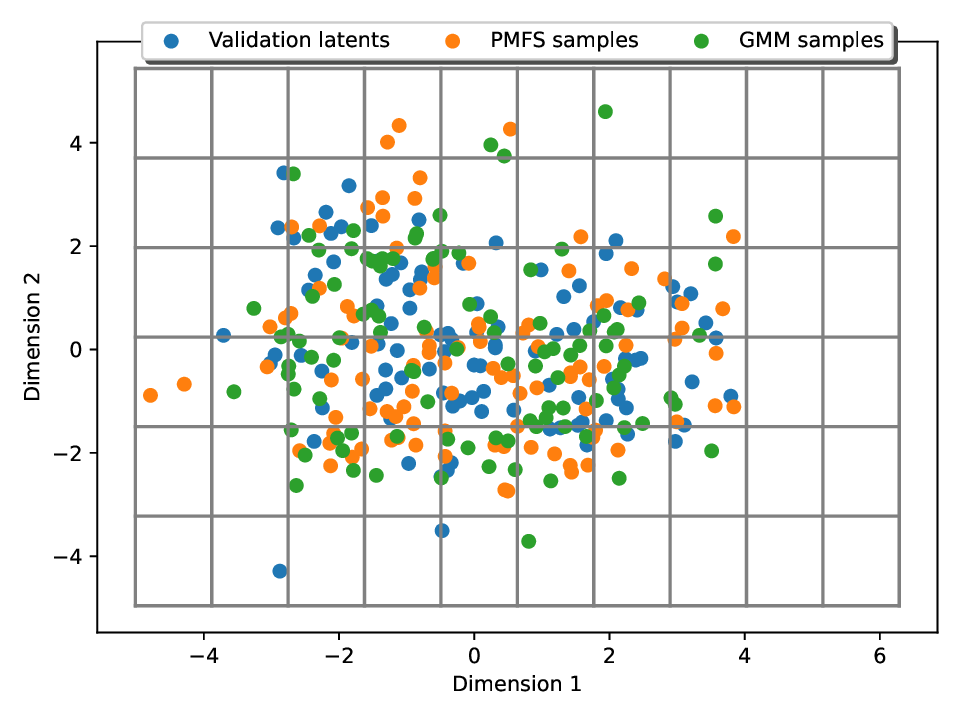}
         \caption{Latent space distribution of VAE.}
     \end{subfigure}
     \hfill
     \begin{subfigure} {0.49\textwidth}
         \centering
         \includegraphics[width=.9\textwidth]{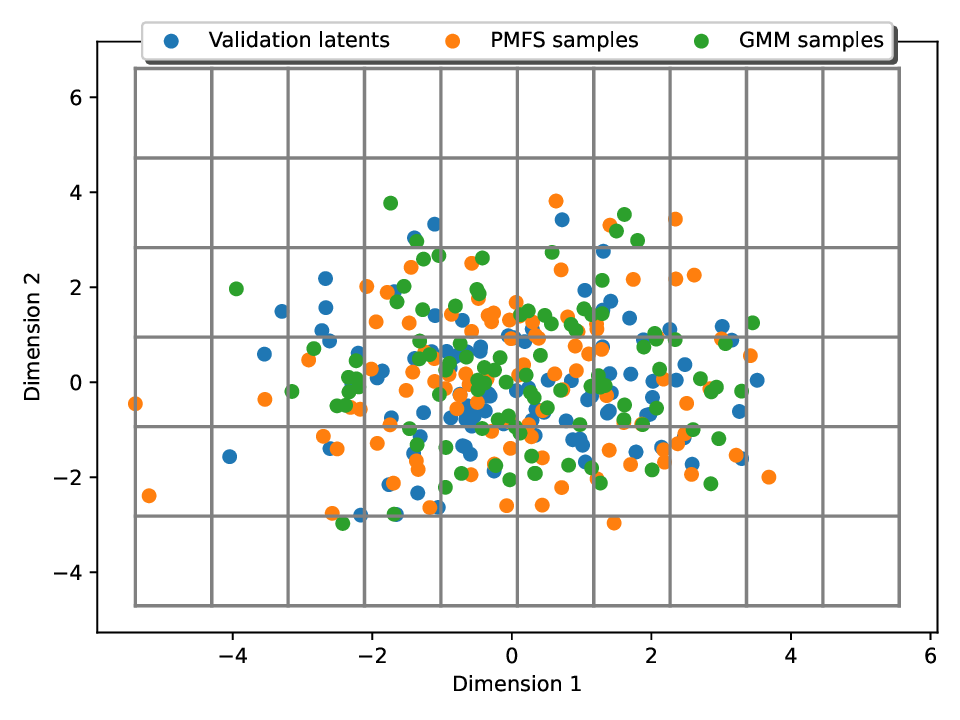}
         \caption{Latent space distribution of $\beta$-VAE.}
     \end{subfigure}
     \hfill
     \begin{subfigure} {0.49\textwidth}
         \centering
         \includegraphics[width=.9\textwidth]{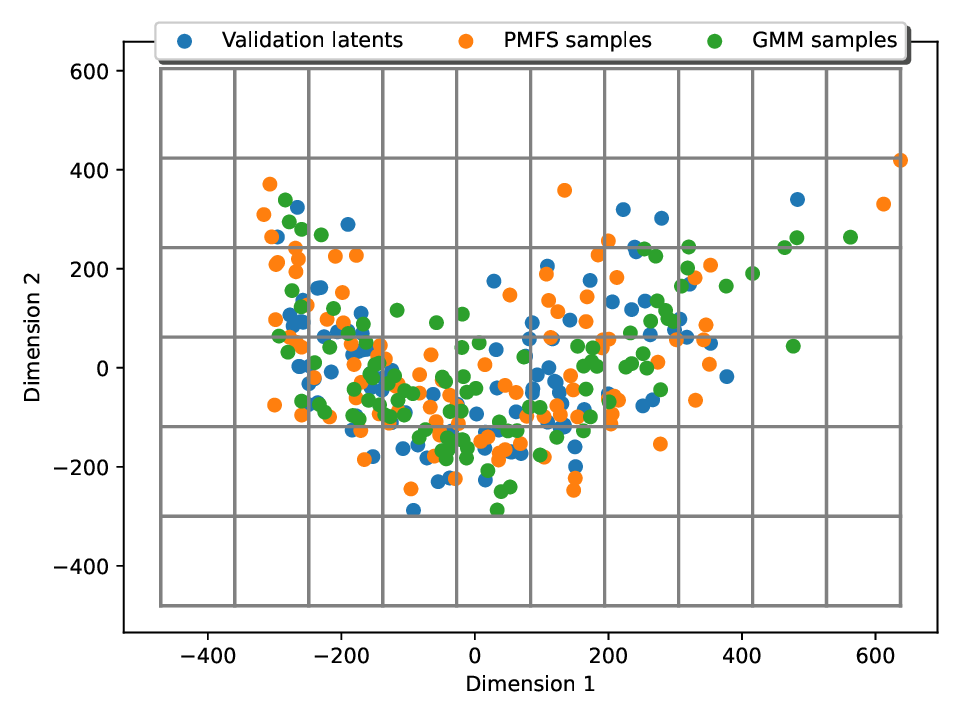}
         \caption{Latent space distribution of WAE.}
     \end{subfigure}
     \hfill
     \begin{subfigure} {0.49\textwidth}
         \centering
         \includegraphics[width=.9\textwidth]{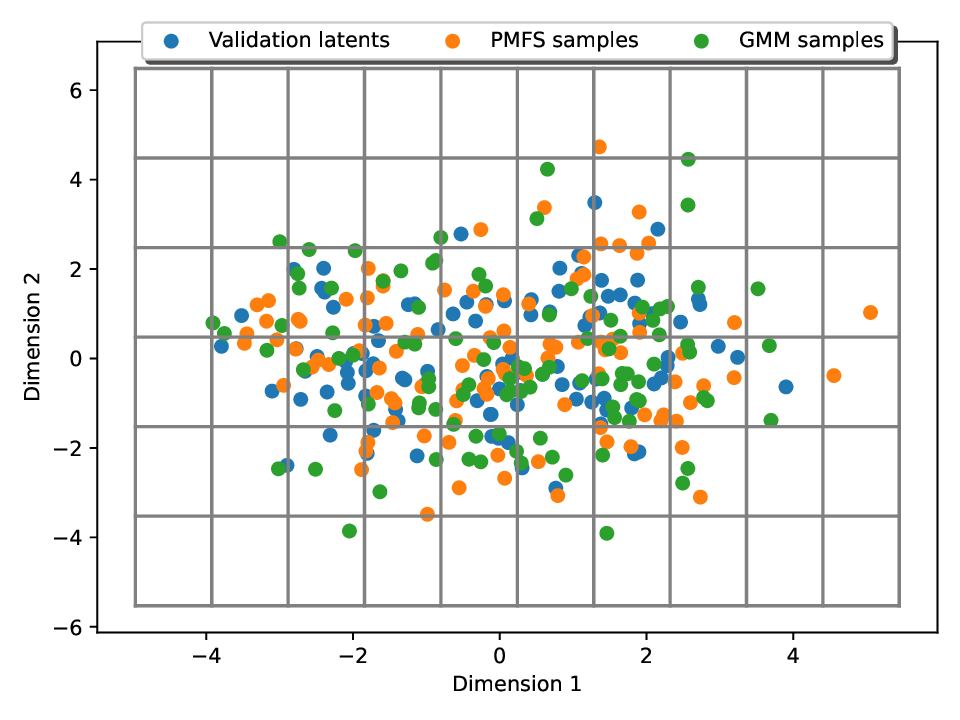}
         \caption{Latent space distribution of InfoVAE.}
     \end{subfigure}
     \hfill
     \caption{Distribution of the validation set’s latent vectors of different Autoencoder models and samples generated via GMM and PMFS sampling on the CelebA data set. These figures represents a dimensionality reduction using PCA into the two dimensional $\mathbb{R}^2$ space of the first hundred images from the validation set, as well as a hundred samples from GMM and PMFS sampling methods.}
\end{figure}

\newpage
\subsection{MOBIUS}
\begin{figure}[H]
    \centering
    \begin{subfigure} {0.49\textwidth}
         \centering
         \includegraphics[width=.9\textwidth]{figures/MOBIUS/latent_dist_ae.eps}
         \caption{Latent space distribution of Vanilla Autoencoder.}
     \end{subfigure}
     \hfill
     \begin{subfigure} {0.49\textwidth}
         \centering
         \includegraphics[width=.9\textwidth]{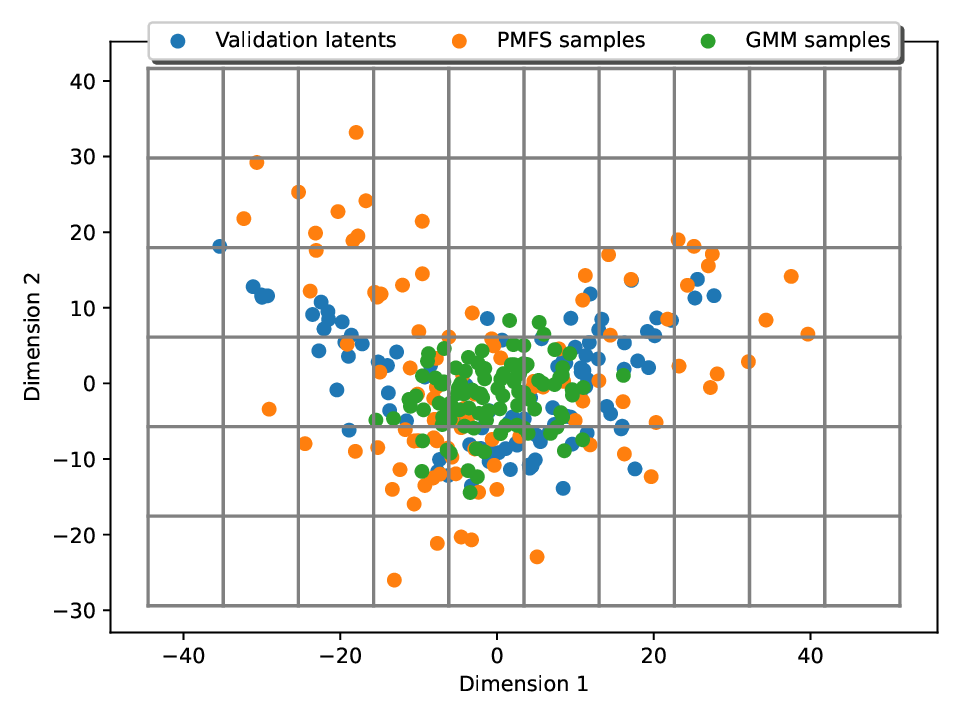}
         \caption{Latent space distribution of VAE.}
     \end{subfigure}
     \hfill
     \begin{subfigure} {0.49\textwidth}
         \centering
         \includegraphics[width=.9\textwidth]{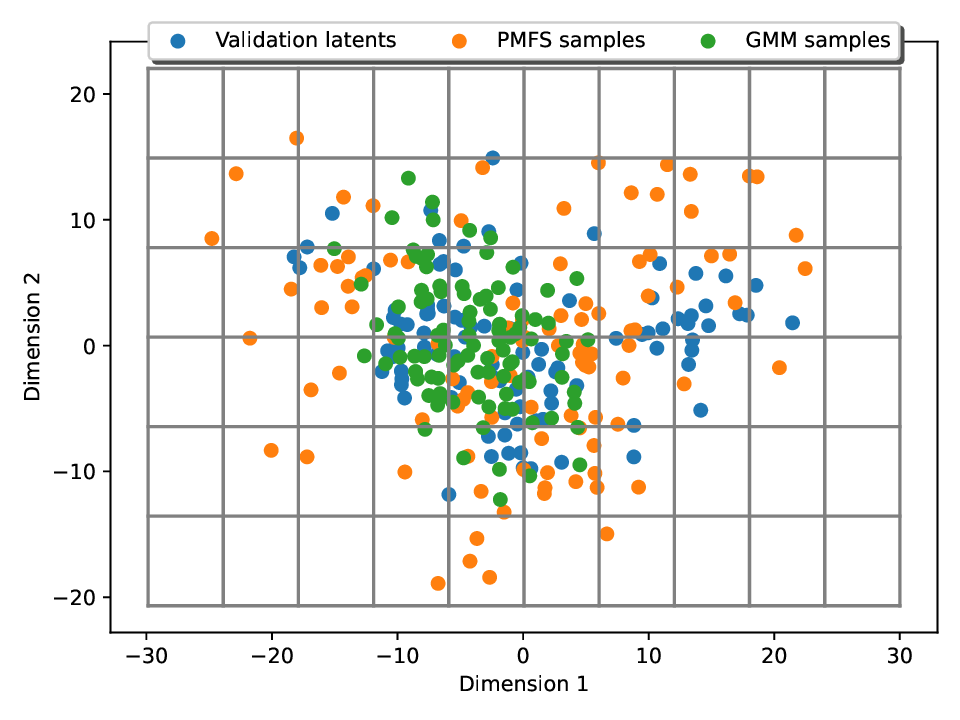}
         \caption{Latent space distribution of $\beta$-VAE.}
     \end{subfigure}
     \hfill
     \begin{subfigure} {0.49\textwidth}
         \centering
         \includegraphics[width=.9\textwidth]{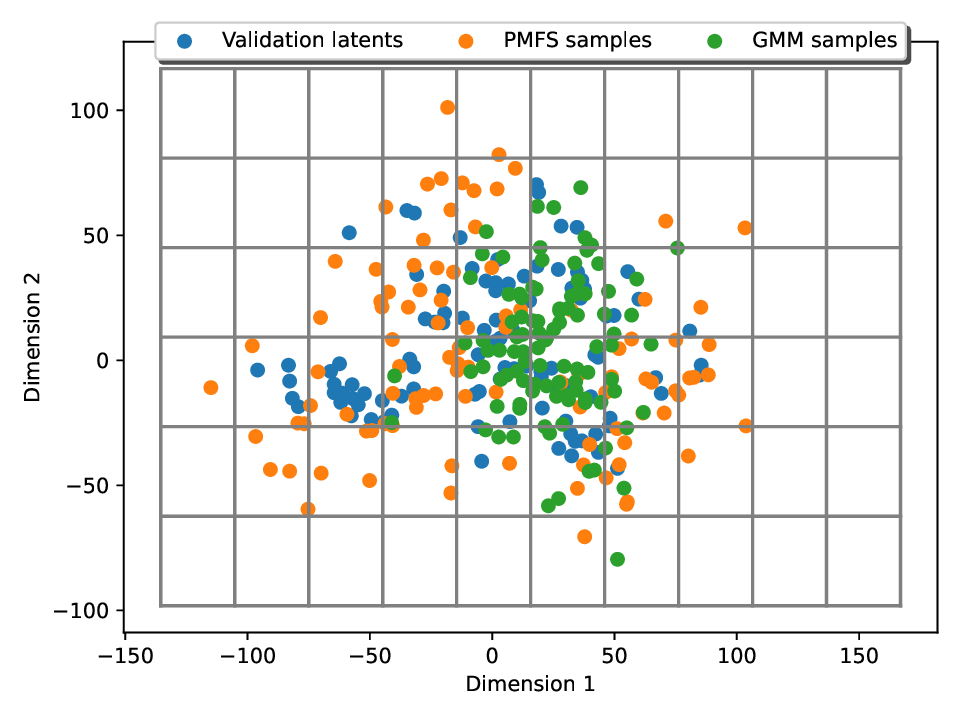}
         \caption{Latent space distribution of WAE.}
     \end{subfigure}
     \hfill
     \begin{subfigure} {0.49\textwidth}
         \centering
         \includegraphics[width=.9\textwidth]{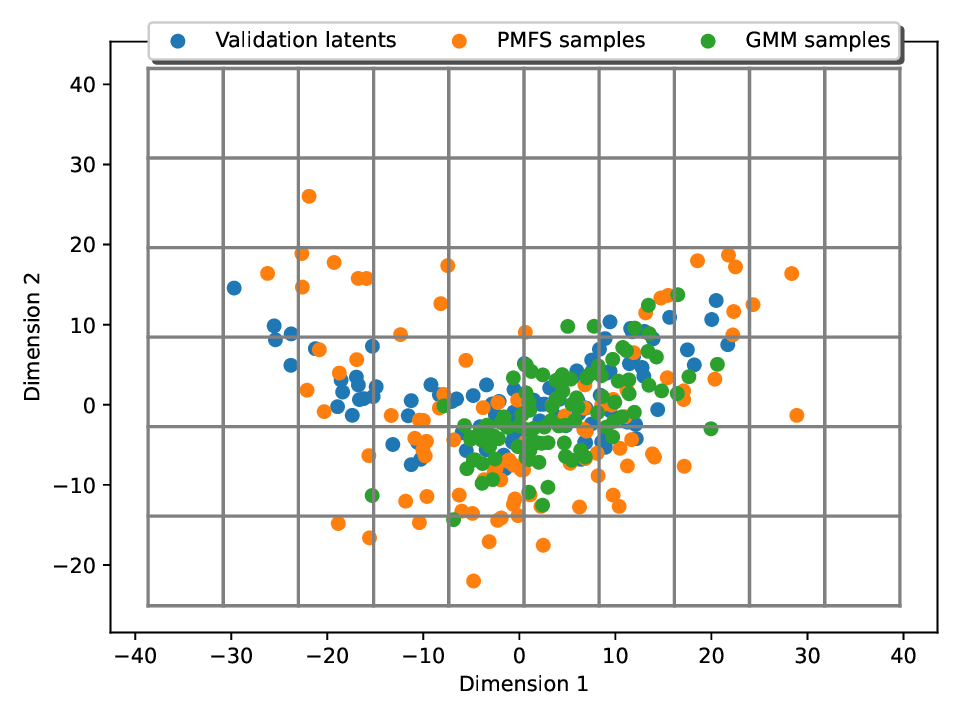}
         \caption{Latent space distribution of InfoVAE.}
     \end{subfigure}
     \hfill
     \caption{Distribution of the validation set’s latent vectors of different Autoencoder models and samples generated via GMM and PMFS sampling on the MOBIUS data set. These figures represents a dimensionality reduction using PCA into the two dimensional $\mathbb{R}^2$ space of the first hundred images from the validation set, as well as a hundred samples from GMM and PMFS sampling methods.}
     \label{dist mobius}
\end{figure}

\end{document}